\documentclass[journal,transmag]{IEEEtran}
\usepackage[backend=bibtex,citestyle=numeric-comp,bibstyle=ieee,sorting=nty,defernumbers=true,giveninits=true,doi=false,isbn=false,url=false,eprint=false]{biblatex}
\usepackage{hyperref}
\usepackage[activate]{microtype}
\ifCLASSINFOpdf
   \usepackage[pdftex]{graphicx}
\else
   \usepackage[dvips]{graphicx}
\fi
\usepackage{amsmath,amssymb}
\usepackage{array}
\usepackage{subcaption}

\newcommand{\ie}{i.\,e., }
\begin{document}

\title{End-to-End Video-To-Speech Synthesis using Generative Adversarial Networks}

\author{\IEEEauthorblockN{Rodrigo Mira,
Konstantinos Vougioukas,
Pingchuan Ma, 
Stavros Petridis~\IEEEmembership{Member,~IEEE}\\ 
Björn W.\ Schuller~\IEEEmembership{Fellow,~IEEE}, 
Maja Pantic~\IEEEmembership{Fellow,~IEEE}} 
\thanks{This work has been submitted to the IEEE for possible publication. Copyright may be transferred without notice, after which this version may no longer be accessible.
Manuscript received Month Day, Year; revised Month Day, Year. 
Corresponding author: R. Mira (email: rs2517@ic.ac.uk). Rodrigo Mira would like to thank Samsung for their continued support of his work on this project.

Rodrigo Mira, Konstantinos Vougioukas, Pingchuan Ma, Stavros Petridis, Björn W.\ Schuller, Maja Pantic are with the IBUG Group, Department of Computing, Imperial College London, UK

Stavros Petridis is with the Samsung AI Centre Cambridge, UK.

Björn W.\ Schuller is with the Chair of Embedded Intelligence for Health Care and Wellbeing, University of Augsburg, Germany.

Maja Pantic is with Facebook London, UK.
}}

\markboth{}
{Shell \MakeLowercase{\textit{et al.}}: Bare Demo of IEEEtran.cls for IEEE Transactions on Magnetics Journals}
\IEEEtitleabstractindextext{
\begin{abstract}

Video-to-speech is the process of reconstructing the audio speech from a video of a spoken utterance. Previous approaches to this task have relied on a two-step process where an intermediate representation is inferred from the video, and is then decoded into waveform audio using a vocoder or a waveform reconstruction algorithm.
In this work, we propose a new end-to-end video-to-speech model based on Generative Adversarial Networks (GANs) which translates spoken video to waveform end-to-end without using any intermediate representation or separate waveform synthesis algorithm.
Our model consists of an encoder-decoder architecture that receives raw video as input and generates speech, which is then fed to a waveform critic and a power critic. The use of an adversarial loss based on these two critics enables the direct synthesis of raw audio waveform and ensures its realism. In addition, the use of our three comparative losses helps establish direct correspondence between the generated audio and the input video. 
We show that this model is able to reconstruct speech with remarkable realism for constrained datasets such as GRID, and that it is the first end-to-end model to produce intelligible speech for LRW (Lip Reading in the Wild), featuring hundreds of speakers recorded entirely `in the wild'. We evaluate the generated samples in two different scenarios -- seen and unseen speakers -- using four objective  metrics which measure the quality and intelligibility of artificial speech. We demonstrate that the proposed approach outperforms all previous works in most metrics on GRID and LRW.  
 \end{abstract}
 }

\maketitle
\IEEEdisplaynontitleabstractindextext
\IEEEpeerreviewmaketitle
\section{Introduction}

\IEEEPARstart{A}{utomatic} speech recognition (ASR) is a well established field with diverse applications including captioning voiced speech and recognizing voice commands. Deep learning has revolutionised this task in the past years, to the point where state of the art models are able to achieve very low word error rates (WER) \cite{DBLP:conf/interspeech/LiLGLKCNG19}. Although these models are reliable for clean audio, they struggle under noisy conditions \cite{DBLP:conf/interspeech/MaasLOVNN12,DBLP:journals/tcyb/StewartSPM14}, and they are not effective when gaps are found in the audio stream \cite{DBLP:conf/iccv/Zhou0X0019}. The recurrence of these edge cases has driven researchers towards Visual Speech Recognition (VSR), also known as lipreading, which performs speech recognition based on video only. 

Although the translation from video-to-text can now be achieved with remarkable consistency, there are various applications that would benefit from a video-to-audio model, such as videoconferencing in noisy conditions; speech inpainting \cite{DBLP:conf/iccv/Zhou0X0019}, \ie filling in audio gaps from video in an audiovisual stream; or generating an artificial voice for people suffering from aphonia (\ie people who are unable to produce voiced sound). One approach for this task would be to simply combine a lipreading model (which outputs text) with a text-to-speech (TTS) model (which outputs audio). This approach is especially attractive since state-of-the-art TTS models can now produce realistic speech with considerable efficacy \cite{DBLP:conf/icassp/ShenPWSJYCZWRSA18,DBLP:conf/icml/OordLBSVKDLCSCG18}. 

Combining video-to-text and text-to-speech models to perform video-to-speech has, however, some disadvantages. Firstly, these models require large transcribed datasets, since they are trained with text supervision. This is a sizeable constraint given that generating transcripts is a time consuming and expensive process. Secondly, generation can only happen as each word is recognized, which imposes a delay on the throughput of the model, jeopardizing the viability of real-time synthesis. Lastly, using text as an intermediate representation removes any intonation and emotion from the spoken statement, which are fundamental for natural sounding speech.

Given these constraints, some authors have developed end-to-end video-to-speech models which circumvent these issues.
The first of these models \cite{DBLP:conf/interspeech/CornuM15} used visual features based on discrete cosine transform (DCT) and active appearance models (AAM) to predict linear predictive coding (LPC) coefficients and mel-filterbank amplitudes.
Following works have mostly focused on predicting spectrograms \cite{DBLP:conf/iccvw/EphratHP17,DBLP:conf/icassp/AkbariACM18,DBLP:journals/corr/abs-2005-08209}, which is also a common practice in text-to-speech works \cite{DBLP:conf/icassp/ShenPWSJYCZWRSA18}. These models achieve intelligible results, but are only applied to seen speakers, \ie there is exact correspondence between the speakers in the training, validation and test sets, or choose to focus on single speaker speech reconstruction \cite{DBLP:journals/corr/abs-2005-08209}. Recently, \cite{DBLP:journals/corr/abs-2004-02541} has proposed an alternative approach based on predicting WORLD vocoder parameters \cite{DBLP:journals/ieicet/MoriseYO16} which generates clear speech for unseen speakers as well.  However, the reconstructed speech is still not realistic.

It is clear that previous works have avoided synthesising raw audio, likely due to the lack of a suitable loss function, and have focused on generating intermediate representations which are then used for reconstructing speech. To the best of our knowledge, the only work which directly synthesises the raw audio waveform from video is \cite{DBLP:journals/corr/abs-1906-06301}. This work introduces the use of 
GANs \cite{DBLP:journals/corr/GoodfellowPMXWOCB14,DBLP:journals/corr/ArjovskyCB17}, and thanks to the adversarial loss, it is able to directly reconstruct the audio waveform.  This approach also produces realistic utterances for seen speakers, and is the first to produce intelligible speech for unseen speakers.

Our work builds upon the model presented in \cite{DBLP:journals/corr/abs-1906-06301} by proposing architectural changes to the model, and to the training procedure. Firstly, we replace the original encoder composed of five stacked convolutional layers with a ResNet-18 \cite{DBLP:conf/cvpr/HeZRS16} composed of a front end 3D convolutional layer (followed by a max pooling layer), four blocks containing four convolutional layers each and an average pooling layer. Additionally, we replace the GRU (Gated Recurrent Unit) layer following the encoder with two bidirectional GRU layers, increasing the capacity of our temporal model. The adversarial methodology was a major factor towards generating intelligible waveform audio in \cite{DBLP:journals/corr/abs-1906-06301}. Hence, our approach is also based on the Wasserstein GAN \cite{DBLP:journals/corr/ArjovskyCB17}, but we propose a new critic adapted from \cite{DBLP:conf/nips/KumarKBGTSBBC19}. We also propose an additional
critic which discriminates real from synthesized spectrograms. 

Furthermore, we revise the loss configuration presented in \cite{DBLP:journals/corr/abs-1906-06301}. Firstly, we decide to forego the use of the total variation loss and the L1 loss, as their benefit was minimal. Secondly, we use the recently proposed PASE (Problem Agnostic Speech Encoder) \cite{DBLP:journals/corr/abs-1904-03416} as a perceptual feature extractor. Finally, we propose two additional losses, the power loss and the MFCC loss. The power loss is an L1 loss between the (log-scaled) spectrograms of the real and generated waveforms. The MFCC loss is an L1 Loss between the MFCCs (Mel Frequency Cepstral Coefficients) of the real and generated waveforms.

Our contributions for this work are described as follows:
\textbf{1)} We propose a new approach for reconstructing waveform speech directly from video based on GANs without using any intermediate representations. We use two separate critics to discriminate real from synthesized waveforms and spectrograms respectively, and apply three comparative losses to improve the quality of outputs.
\textbf{2)} We include a detailed ablation study where we measure the effect of each component on the final model. We also investigate how the type of visual input, size of training set and range of vocabulary affect the performance.
\textbf{3)} We show results on two different datasets (GRID \cite{grid} and TCD-TIMIT \cite{7050271}) for seen speakers. We find that our model substantially outperforms the state-of-the-art for GRID and adapts well to a larger pool of speakers.
\textbf{4)} We also include results for unseen speakers on two datasets (GRID and LRW \cite{Chung16}). We show that our model achieves intelligible results, even when applied to utterances recorded `in the wild', and outperforms the state-of-the-art for the corpora we present.
\textbf{5)} Finally, we study our model's ability to generalize for videos of silent speakers, and discuss our findings.
    
\section{Related Work} \label{section2}
Video-driven speech reconstruction is effectively the combination of two tasks: lipreading and speech synthesis. As such, we begin by briefly describing the main works in each field, and then go on to describe existing approaches for video-to-speech.
\subsection{Lipreading}
Traditional lipreading approaches relied on  HMMs (Hidden Markov Models) \cite{DBLP:journals/tsp/GurbanT09} or SVMs (Support Vector Machines) \cite{DBLP:journals/tmm/ZhaoBP09} to transcribe videos from manually extracted features such as DCT \cite{DBLP:journals/tsp/GurbanT09} or mouth geometry \cite{DBLP:conf/icassp/KumarCS07}. Recently, end-to-end models have attracted attention due to their superior performance over traditional approaches. One of the first end-to-end architectures for lipreading was \cite{DBLP:journals/corr/AssaelSWF16}. This model featured a convolutional encoder as the visual feature extractor and a two-layer BGRU-RNN (Bidirectional GRU recurrent neural network) followed by a linear layer as the classifier, and it achieved state of the art performance for the GRID corpus. This work was followed by \cite{DBLP:conf/accv/ChungZ16}, whose model relied entirely on CNNs (Convolutional Neural Networks) and RNNs, and  was successfully applied to spoken utterances recorded in the wild.

Various works have followed which apply end-to-end deep learning models to achieve competitive lipreading performance. \cite{DBLP:conf/icassp/PetridisLP17,DBLP:journals/prl/PetridisWMLP20} propose an encoder composed of fully connected layers and performs classification using LSTMs (Long-short Term Memory RNNs). Other works choose to use convolutional encoders \cite{DBLP:conf/interspeech/ShillingfordAHP19}, often featuring residual connections \cite{DBLP:conf/interspeech/StafylakisT17}, and then apply RNNs to perform classification. Furthermore, these end-to-end architectures have been extended for multi-view lipreading \cite{DBLP:conf/bmvc/PetridisWLP17} and audiovisual \cite{DBLP:conf/icassp/PetridisSMCTP18} speech recognition.  

\subsection{Speech Synthesis} \label{wss}
One of the most popular speech synthesis models in recent years has been WaveNet \cite{DBLP:conf/ssw/OordDZSVGKSK16}, which proposed dilated causal convolutions to compose waveform audio sample by sample, taking advantage of the large receptive field achieved by stacking these layers. This model achieved far more realistic results than any artificial synthesizer proposed before then. Another work \cite{DBLP:conf/interspeech/WangSSWWJYXCBLA17} introduced a vastly different sequence-to-sequence model that predicted linear-scale spectrograms from text, which were then converted into waveform using the Griffin-Lim Algorithm (GLA) \cite{DBLP:journals/taslp/ZhuBW07}. This process produced very clear and intelligible audio. In the following years, \cite{DBLP:conf/icassp/ShenPWSJYCZWRSA18} combined these two methodologies to push the state-of-the-art once more, and \cite{DBLP:conf/icml/OordLBSVKDLCSCG18} accelerated and improved the original WaveNet.

The first model to apply GANs for end-to-end speech synthesis was \cite{DBLP:conf/iclr/DonahueMP19}, which used simple convolutional networks with large kernels as the generator and discriminator and applied the improved Wasserstein loss \cite{DBLP:conf/nips/GulrajaniAADC17}. 
In a later work \cite{DBLP:journals/corr/abs-1910-11480}, the original WaveNet vocoder \cite{DBLP:conf/ssw/OordDZSVGKSK16} has been combined with the adversarial methodology introduced in \cite{DBLP:conf/iclr/DonahueMP19}. This results in a network which has far less  parameters than the original WaveNet, but remains on par with the latest WaveNet-based models. Recently, the first end-to-end adversarial Text-To-Speech model \cite{DBLP:journals/corr/abs-2006-03575} was also proposed, whose performance is comparable to the state-of-the-art. 

\subsection{Reconstructing audio from visual speech}
To the best of the authors' knowledge, the first work to attempt the task of video-to-speech synthesis directly was \cite{DBLP:conf/interspeech/CornuM15}. The proposed model aims to predict the spectral envelope (LPC or mel-filterbanks) from manually extracted visual features (DCT or AAM) using Gaussian Mixture Models (GMMs) or deep neural networks. These acoustic parameters are then fed into an HMM-based vocoder, together with an estimate of the voicing parameters. Through multiple user studies, the speech reconstructed by this model is shown to have fairly low intelligibility (WER $\approx 50\,\%$), but shows that this task is indeed achievable. This work was extended in \cite{DBLP:journals/taslp/CornuM17}, which introduced additional temporal information in the visual features and in the model itself. These improvements yielded an impressive 15\,\% WER for GRID (single speaker), based on user studies.

The next development in this field comes with \cite{DBLP:conf/icassp/EphratP17}, which uses a deep CNN architecture to predict acoustic features -- LPC analysis followed by LSP (Line Spectral Pairs) decomposition, frame by frame --  from gray-scale video frames. These are combined with white noise (excitation signal) and fed into a source-filter speech synthesizer which produces unvoiced speech. This model produces intelligible results (WER $< 20\,\%$) when trained and tested on a single speaker from GRID, and constitutes a step forward given that it no longer relies on handcrafted visual features as input. An improved version of this model was presented in \cite{DBLP:conf/iccvw/EphratHP17}, which predicts spectrograms that are then translated into waveform using the Griffin-Lim algorithm. This extension also proposes a new encoder composed of two ResNet-18s followed by a post-processing network which increases temporal resolution. This work is the first to experiment with multiple speakers and achieves much more realistic speech than any previous work for this task.

Lip2Audspec \cite{DBLP:conf/icassp/AkbariACM18} proposes a similar CNN+RNN encoder to predict spectrograms directly from the gray-scale frames of the video. As in \cite{DBLP:conf/iccvw/EphratHP17}, the spectrograms are converted to waveform using a phase estimation method. The resulting spectrograms are very close to the original samples, but the reconstructed waveforms sound noticeably robotic. Another recent work \cite{DBLP:journals/corr/abs-2004-02541} uses CNNs+RNNs to predict vocoder parameters (aperiodicity and spectral envelope), rather than spectrograms. Additionally, the model is trained to predict the transcription of the speech, in other words performing speech reconstruction and recognition simultaneously in a multi-task fashion. This approach achieves results which are very impressive when measured with objective speech quality metrics (PESQ, STOI), but yields samples which still sound noticeably robotic. 

Finally, a recent work \cite{DBLP:journals/corr/abs-2005-08209} proposes an approach based on the Tacotron 2 architecture \cite{DBLP:conf/icassp/ShenPWSJYCZWRSA18}, predicting mel-frequency spectrograms from video rather than text. To perform this task, it applies a stack of residual 3D convolutional layers as a spatio-temporal encoder for the video, and combines it with an attention-based decoder adapted from \cite{DBLP:conf/icassp/ShenPWSJYCZWRSA18}, which generates the spectrograms. Unlike Tacotron, these spectrograms are decoded into waveform audio using the Griffin-Lim algorithm \cite{DBLP:journals/taslp/ZhuBW07} rather than WaveNet, as the authors claim the generated spectrograms are not as accurate as modern TTS works, and therefore do not perform well with neural vocoders. This work is able to generate remarkably intelligible audio from visual speech and achieves state-of-the-art performance in all presented metrics. However, it focuses on speaker specific speech reconstruction, 

\ie it is trained and tested on the same speaker.

An aspect which is worth highlighting is that none of these models attempt to generate the waveform end-to-end from video, instead predicting spectrograms or other features which can be translated into waveform. This is likely due to the notoriously arduous task of generating realistic waveforms, which can be attributed to the lack of suitable loss functions. The only model to perform video-to-waveform speech reconstruction without the use of intermediate representations is \cite{DBLP:journals/corr/abs-1906-06301}. This work proposes a generative adversarial network based on a convolutional encoder-decoder model (combined with a GRU) which encodes video into visual features and decodes them directly into waveform audio. The generator is trained with an adversarial loss based on a convolutional waveform critic, as well as three other comparative losses. This procedure achieves competitive results for speech reconstruction on both seen and unseen speaker datasets (GRID).  

\subsection{Reconstructing audio from multi-view visual speech}
The majority of works in video-driven speech reconstruction use frontal views of the face. In this section, we briefly describe a set of works which use multiple-views in order to improve the quality of reconstructed speech. 

The first work to use multi-view video for this task was \cite{DBLP:conf/mm/KumarANSSZ18}. This model is very similar to \cite{DBLP:conf/icassp/EphratP17} in the sense that it applies a CNN to extract visual features directly from video, which then predict vocoder parameters (LPC followed by LSP). This work, however, uses video taken from two different angles for every speaker (Oulu VS2 dataset \cite{DBLP:conf/fgr/AninaZZP15}). The results presented in this paper show that the use of multiple views can substantially improve speech reconstruction performance.

This model has been improved in \cite{DBLP:conf/ism/KumarJSSZY18} by replacing the LSTM with a BGRU and using more than two views as input. It is shown that the use of three views can yield improvements of 20\,\% in the quality of reconstructed outputs (measured with PESQ). This has been extended in \cite{DBLP:conf/aaai/KumarJSSYZ19} by including a view classifier to attribute view labels to the input videos and by also generating text transcriptions. The latest work in this field \cite{DBLP:conf/interspeech/UttamKSASMS19} follows the trend seen in single-view speech reconstruction research \cite{DBLP:conf/icassp/EphratP17,DBLP:conf/iccvw/EphratHP17} and speech synthesis in general \cite{5931414,DBLP:conf/interspeech/WangSSWWJYXCBLA17} by switching from LPC coefficients to spectrograms as the predicted audio representation. 

\subsection{Audio reconstruction from video in other applications }
Finally, a set of past works has approached the application of Video-to-Audio models to domains outside speech \cite{DBLP:conf/cvpr/OwensIMTAF16,DBLP:conf/cvpr/ZhouWFBB18,DBLP:conf/mm/ChenSDX17}. Namely, these papers have focused on a diverse range of datasets which feature a set of generic sounds such as fireworks and drums \cite{DBLP:conf/cvpr/ZhouWFBB18}; different instruments being played \cite{DBLP:conf/mm/ChenSDX17}; or even objects composed of different materials being hit with a drumstick \cite{DBLP:conf/cvpr/OwensIMTAF16}. The methodology applied to reconstruct audio from video is similar to what is seen for video-to-speech systems. CNNs are applied to encode the video frames, followed by RNNs or fully connected layers to produce acoustic features which are decoded into audio using vocoders. While some of these works struggle to reproduce the corresponding audio, \cite{DBLP:conf/cvpr/ZhouWFBB18} is able to produce remarkably realistic audio (as proven by its user studies) by combining the extraction of optical flow with a neural network-based vocoder.

\section{Video-Driven Speech Reconstruction} \label{section3}
\begin{figure}
\centering 
\includegraphics[width = 0.86\linewidth]{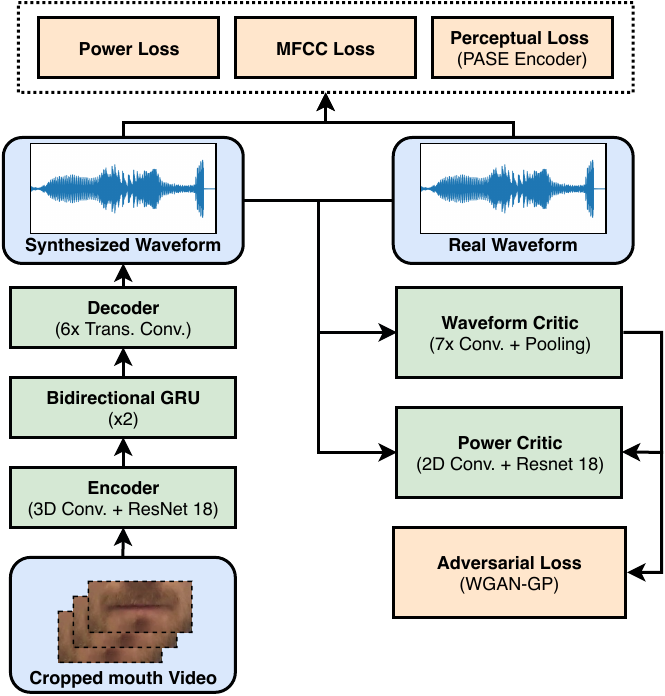}
\caption{Architecture of the generator (encoder, bidirectional GRU, decoder) and critics (waveform critic, power critic) used in this work, as well as the losses that are used for training.}
\label{architecture}
\end{figure}

Our model is composed of a video encoder based on a ResNet-18 combined with a Bidirectional GRU, as well as a convolutional decoder which transforms the visual features into waveform audio. This generator is trained using two separate critics, to ensure the realism of the outputs, as well as three L1 losses to minimize the difference between real and synthesized audio for each video.

\subsection{Generator}
\begin{figure*}
\centering 
\includegraphics[width = 0.9\linewidth]{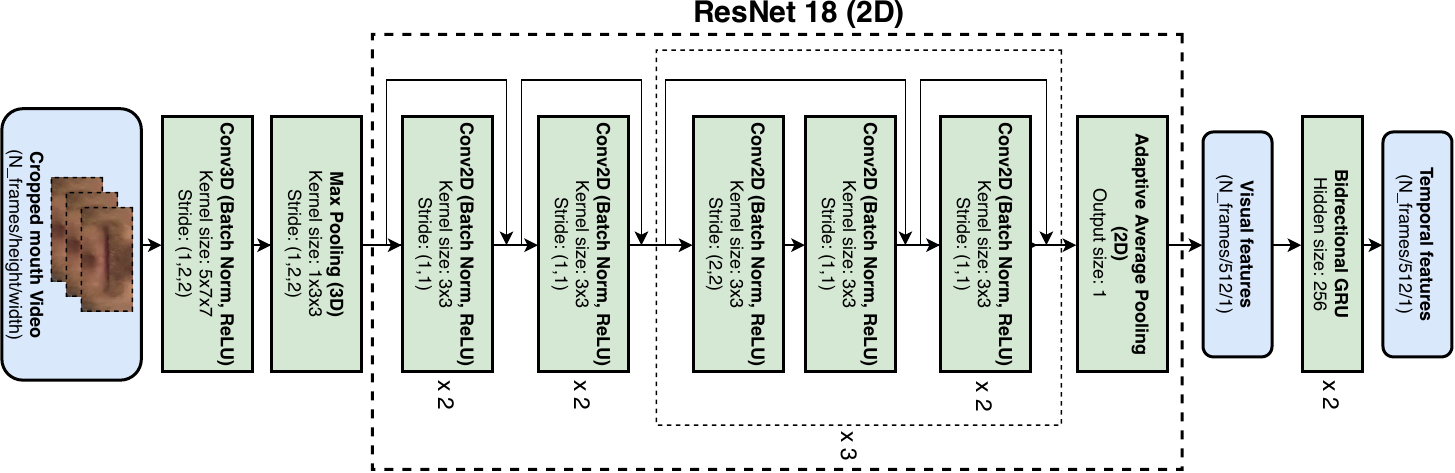}
\caption{Description of the layers in the encoder (generator).}
\label{encoder}
\end{figure*}

Given that we aim to synthesize speech directly from video, our generator accomplishes two sequential tasks: encode temporal visual features and decode them into an audio waveform. Firstly, we encode the frames of the video using a Resnet-18 preceded by a spatio-temporal 3D convolutional layer (combined with a max pooling layer). This initial layer has a receptive field of 5 frames centered on the frame it will encode meaning that the encoding for each frame will depend on the previous two frames and on the following two frames. We experimented with different numbers of frames as input to this layer (3 and 7), but found that this did not considerably affect results. The ResNet-18 is composed of 4 blocks of 4 convolutional layers, each followed by batch normalization and ReLU (Rectified Linear Unit) activation, and an adaptive average pooling layer. The features extracted from the ResNet encoder are then fed into a 2-layer bidirectional GRU which temporally correlates the features produced from each set of frames. This architecture is described in detail in Figure \ref{encoder}.

\begin{figure}
\centering 
\includegraphics[width = 0.9\linewidth]{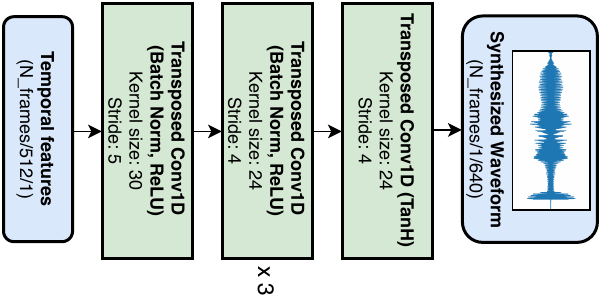}
\caption{Description of the layers in the decoder (generator).}
\label{decoder}
\end{figure}

After this, the decoder upsamples the features from each video frame into a waveform segment of $N$ audio samples. The length of each segment is given by:
\begin{align}
    N = \frac{audio\ sampling\ rate}{video\ frame\ rate}.
\end{align}
Since we use a sampling rate of 16\,kHz and a frame rate of 25 frames per second, $N$ is equal to 640 (corresponding to 40\,ms of audio). The decoder is composed of six stacked transposed convolutional layers, each followed by batch normalization and ReLU activation except for the last layer which uses a hyperbolic tangent activation function. In an attempt to alleviate the issue of abrupt frame transitions, we use an overlap of 50\,\% between the generated waveform frames, as proposed in \cite{DBLP:conf/icassp/EphratP17}. The overlapped segments are linearly averaged sample by sample in order to maintain the original waveform scale. The detailed architecture of the decoder is shown in Figure \ref{decoder}. 

\subsection{Critics}
\begin{figure}
\centering 
\includegraphics[width = 0.9\linewidth]{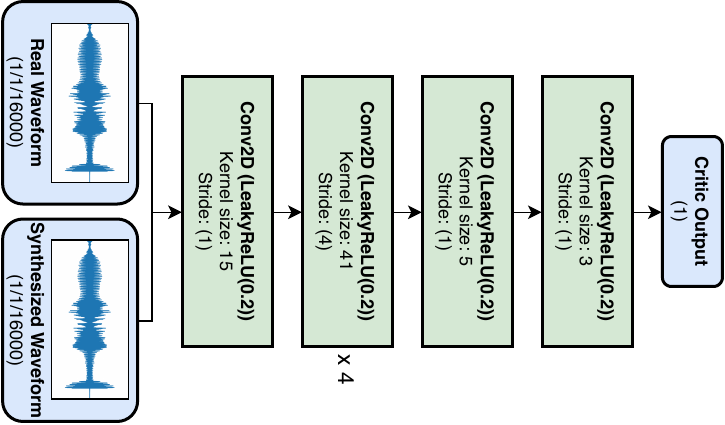}
\caption{Description of the layers in the waveform critic used to train our model.}
\label{wave_critic}
\end{figure}

As demonstrated in recent works \cite{DBLP:conf/iclr/DonahueMP19,DBLP:conf/nips/KumarKBGTSBBC19,DBLP:journals/corr/abs-1910-11480}, the use of a waveform critic can dramatically increase the realism and clarity of synthesized speech. To discriminate the real from the synthesized waveforms, we adapt the critic from \cite{DBLP:conf/nips/KumarKBGTSBBC19}. After experimenting extensively with and without weight normalization for this module, as well as for the generator, we find that weight normalization increases the stability of adversarial training but overall leads to worse results. Therefore, we remove weight normalization from this critic but otherwise keep the original architecture: 7 convolutional layers, each followed by Leaky ReLU activation, as shown in Figure \ref{wave_critic}.

We did not attempt batch normalization, which worked well for the generator, since this interferes with the gradient penalty for our adversarial loss \cite{DBLP:conf/nips/GulrajaniAADC17}. We compared this architecture to other convolutional critics similar to the one proposed in \cite{DBLP:conf/iclr/DonahueMP19} as well as a one-dimensional ResNet 18, and found that this critic produced the best results. Remarkably, this critic has a far smaller receptive field than any of the critics we experimented with. This may indicate that waveform critics work best when focusing on the small scale.

\begin{figure}
\centering 
\includegraphics[width = 0.9\linewidth]{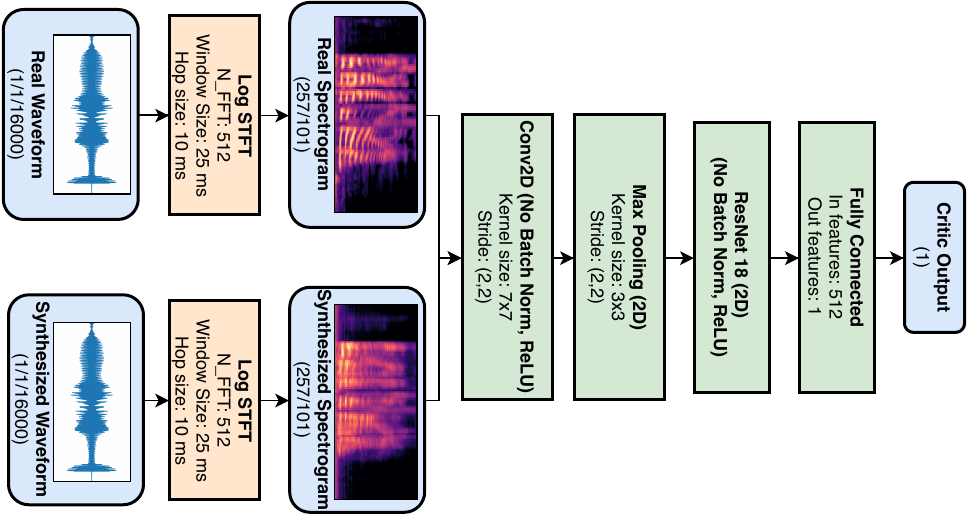}
\caption{Description of the layers in the power critic used to train our model, and the process used to extract spectrograms from waveform samples.}
\label{power_critic}
\end{figure}

Inspired by the SpecGAN model \cite{DBLP:conf/iclr/DonahueMP19}, we propose to combine the waveform critic, which judges the audio in the temporal domain, with a power critic, which judges the audio in the spectral domain. This module discriminates the spectrograms computed from real and generated audio. We first compute the spectrogram from both the real and generated samples using the short-time Fourier transform (STFT) with a window size of 25\,ms, a hop size of 10\,ms and frequency bins of size 512. We then compute the natural logarithm of the spectrogram magnitudes, normalize these values to mean 0 and variance 1, clip values outside [-3,3] and normalize them to [-1,1], similarly to \cite{DBLP:conf/iclr/DonahueMP19}. In this case, we use a ResNet18 identical to the one presented in our generator, except with a two-dimensional front end convolutional layer in the beginning, since our input is a single image. As with the waveform critic, we cannot use batch normalization in this module due to the gradient penalty, and found that weight normalization did not improve results. The architecture for the power critic is shown in Figure \ref{power_critic}.

\subsection{Losses}
To train our network, we apply the Wasserstein GAN loss \cite{DBLP:journals/corr/ArjovskyCB17}, which aims to minimize the Wasserstein Distance between the distributions of real and synthesized data. We also add the gradient penalty \cite{DBLP:conf/nips/GulrajaniAADC17} in order to satisfy the Lipschitz constraint in the Wasserstein GAN objective. The losses for the generator and respective critic(s) are defined as:
\begin{align}\label{eq:1}
    &L_{G} = -\underset{\widetilde{x} \sim \mathbb{P}_{G}}{\mathbb{E}}[D(\widetilde{x})] + \lambda \underset{\hat{x}\sim \mathbb{P}_{\hat{x}}}{\mathbb{E}} [(\|\nabla_{\hat{x}} D(\hat{x})\| - 1)^2]
    \\
    &L_{D} =  \underset{\widetilde{x} \sim \mathbb{P}_{G}}{\mathbb{E}}[D(\widetilde{x})] - \underset{x \sim \mathbb{P}_{R}}{\mathbb{E}}[D(x)],
\end{align}
where $G$ is the generator, $D$ is the critic, $x \sim \mathbb{P}_{R}$ are samples from the real distribution, $\widetilde{x} \sim \mathbb{P}_{G}$ are samples from the estimated distribution (produced by the generator) and $\hat{x}\sim \mathbb{P}_{\hat{x}}$ are sampled uniformly between two points from $P_G$ and $P_R$ respectively. In this work, we apply two critics: the waveform critic and the power critic. Each critic is trained with their own losses $L_{D_{wave}}$ and $L_{D_{power}}$, whereas the generator combines the losses from the two critics such that:
\begin{align}
    &L_{G_{adv}} = L_{G_{wave}} + L_{G_{power}}, 
\end{align}
where $L_{G_{wave}}$ and $L_{G_{power}}$ are calculated as mentioned in Eq. \ref{eq:1}. The coefficient for the gradient penalty $\lambda$ is kept at the value of 10 for both critics, as proposed in \cite{DBLP:conf/nips/GulrajaniAADC17}.

In addition to this adversarial loss, we also apply three other losses to train the generator. The first is a perceptual loss: 
\begin{align}
    L_{PASE} = \| \delta(x) - \delta(\widetilde{x}) \|,
\end{align}
where $x$ is the real waveform, $\widetilde{x}$ is the synthesized waveform from the same video and $\delta$ is our perceptual feature extractor. In this work, we use the pre-trained PASE model \cite{DBLP:journals/corr/abs-1904-03416} to extract perceptual features $\delta(x)$. PASE has been trained in a self-supervised manner to produce meaningful speech representations. We have also tried using PASE+ \cite{DBLP:journals/corr/abs-2001-09239}, which is an improved version of PASE, however, no improvement in the speech reconstruction quality was observed. Furthermore, we experimented with multiple ASR models as feature extractors, but we found that they also did not improve results.

The second loss we apply is the Power Loss. This function aims to improve the accuracy of the reconstructed audio by attempting to match it with the real audio in the frequency domain. For this purpose, we use the L1 loss between the STFT magnitudes of the real and synthesized audio as follows:
\begin{align}
    L_{power} = \| log \|STFT(x)\|^2 - log \|STFT(\widetilde{x})\|^2 \|,
\end{align}
where $x$ is the real waveform, $\widetilde{x}$ is the synthesized waveform from the same video and $STFT$ is the Short Time Fourier Transform with a window size of 25\,ms, a hop size of 10\,ms and frequency bins of size 512 (same parameters used for the power critic). We found that scaling the magnitudes using the natural logarithm and using an L1 Loss rather than the L2 Loss chosen in \cite{DBLP:conf/icml/OordLBSVKDLCSCG18} greatly improve training stability and performance.

The third loss we apply is the MFCC Loss:
\begin{align}
    L_{MFCC} = \| MFCC(x) - MFCC(\widetilde{x}) \|,
\end{align}
where $x$ is the real waveform, $\widetilde{x}$ is the synthesized waveform from the same video and $MFCC$ is the MFCC function which extracts 25 mel-frequency cepstral coefficients from the corresponding waveform. The objective of this loss lies in increasing the accuracy and intelligibility of the synthesized speech, given that MFCCs are known to be effective in ASR \cite{DBLP:conf/iscas/HanCCP06} and emotion recognition \cite{6514336}. 

We adapt the function provided on an open-source repository\footnote{https://github.com/skaws2003/pytorch-mfcc}. 

Finally, the loss for the generator is described based on the losses mentioned above as:
\begin{align}
    L_{G} = \alpha_1 L_{G_{adv}} + \alpha_2 L_{PASE} + \alpha_3 L_{power} + \alpha_4 L_{MFCC}.
\end{align}
We tune the coefficients $\alpha_{1,2,3,4}$ by sequentially training multiple models on GRID (4 speakers, seen speaker split) and incrementally finding the coefficients that yield the best WER on the validation set. Through our search, we find that $\alpha_1 = 1$, $\alpha_2 = 140$, $\alpha_3 = 50$, $\alpha_4 = 0.4$ yield the best results.


\subsection{Training details}
We use the Adam optimizer with a learning rate of 0.0001 and $\beta_1 = 0.5$, $\beta_2 = 0.99 $ to train our generator and critics end-to-end. Given that the critics should be trained to completion before every generator training step, we perform 6 training steps on the critics before every training step of the generator. It should also be noted that we feed a one second clip randomly sampled from the real and synthesized audio to each of the critics, rather than the entire utterance. The other losses are computed using the entire real and synthesized utterances.

Additionally, we employ two data augmentation methods during training. Firstly, we apply random cropping on the input frame, producing a frame with roughly 90\,\% of the original size. Furthermore, we apply horizontal flipping to each frame with a probability of 50\,\%. These procedures help make our model more robust and provide regularization. During test time, the same cropping is performed on the center of the frame and no horizontal flipping is performed.

Training our model for each of the experiments generally takes aprroximately one week on an Nvidia RTX 2080 Ti GPU. Synthesizing a 3 second audio clip sampled at 16\,kHz from 75 frames of video takes approximately 32\,ms on the same high-end GPU, excluding pre-processing.

\section{Datasets} \label{section4}
\begin{table}
\centering 
\def\arraystretch{1.5}
\begin{tabular}{  m{2.2cm} | >{\centering}m{1.9cm}  >{\centering}m{1.6cm}  >{\centering\arraybackslash}m{1.6cm} } 
\hline
Corpus & Training set (clips / hours) & Validation set (clips / hours) & Test set \break (clips / hours) \\
\hline
\hline
GRID (4 speakers, seen speakers) & 3576 / 2.98 & 210 / 0.18 & 210 / 0.18 \\
\hline
GRID (33 speakers, seen speakers) & 29584 / 24.65 & 1642 / 1.37 & 1641 / 1.37 \\
\hline
GRID (33 speakers, unseen speakers) & 15888 / 13.24 & 7000 / 5.83 & 9982 / 8.32 \\
\hline
TCD-TIMIT \hfill \break (3 lipspeakers) & 1014 / 1.64 & 57 / 0.09 & 60 / 0.09 \\
\hline
LRW (full) & 488763 / 157.49 & 25000 / 8.06 & 25000 / 8.06 \\
\hline
FLRW 500 Words & 112811 / 36.35 & 5878 / 1.89 & 5987 / 1.93 \\
\hline
FLRW 100 Words & 22055 / 7.11 & 1151 / 0.37 & 1144 / 0.37 \\
\hline
FLRW 20 Words & 4347 / 1.40 & 266 / 0.09 & 248 / 0.08 \\
\hline
\end{tabular}
\caption{Number of speech clips and total number of hours of speech for each dataset used in our study.}
\label{datasets}
\end{table}
For the purpose of this work, we use three separate audiovisual datasets to train and evaluate our model: GRID, TCD-TIMIT and LRW. GRID contains 33 speakers, each uttering 1000 short sentences composed of 6 simple words from a constrained vocabulary of 51 words. It is the most commonly used dataset for video-driven speech reconstruction \cite{DBLP:conf/icassp/EphratP17,DBLP:conf/icassp/AkbariACM18,DBLP:journals/corr/abs-2004-02541,DBLP:conf/icassp/AkbariACM18,DBLP:journals/corr/abs-2005-08209} due to the clean recording conditions and the limited vocabulary.

TCD-TIMIT is another audiovisual dataset composed of 62 speakers, three of which are trained lipspeakers. In order to compare with previous works \cite{DBLP:journals/corr/abs-2005-08209}, we only use the audiovisual data uttered by the three lipspeakers. Each lipspeaker utters 375 unique phonetically rich sentences, as well as two additional sentences which are uttered by all three speakers. This results in a total of 1\,131 clips. The video/audio for this data is recorded in studio conditions with exceptional clarity given the particular speaking ability of the professional lipspeakers. 

Finally, LRW contains roughly 500\,000 speech samples (500 words, up to 1\,000 clips per word) uttered by hundreds of different speakers, taken from television broadcasts. Due to the fact that these utterances are recorded `in the wild' from a large variety of speakers, LRW presents a far more substantial challenge for speech reconstruction than the datasets mentioned above. Additionally, we use a subset of this corpus which keeps only the videos that are approximately frontal, \ie videos with yaw, pitch and roll below 10 degrees. This leads to a corpus containing 124\,676 samples in total and will be referred to as \textit{F(rontal)LRW}. We also randomly select 20/100 words from this subset to experiment with different ranges of vocabulary during training/testing. These smaller sets will be referred to as \textit{FLRW20} and \textit{FLRW100}, respectively. Further statistics for each dataset are presented in Table \ref{datasets}.

Rather than using the full face as input to our network, as is standard in other speech reconstruction works \cite{DBLP:conf/iccvw/EphratHP17,DBLP:conf/icassp/AkbariACM18,DBLP:journals/corr/abs-2005-08209}, we crop the mouth of the speaker, and use it as the input for every frame. We do this by performing face detection and alignment using dlib's 68 landmark model \cite{dlib09}, aligning each face to a reference mean face shape and extracting a mouth ROI (Region of Interest) from each frame. The mouth ROI is of size 128x74 for GRID and 96x96 for TCD-TIMIT and LRW. 

\section{Evaluation Metrics} \label{section5}
Although many metrics have been proposed for evaluating the quality of speech \cite{DBLP:series/sci/Loizou11}, it is widely acknowledged that none of the existing metrics are highly correlated with human perception. For this reason, we evaluate our speech reconstruction model using 4 objective metrics which capture different properties of the audio: PESQ, STOI, MCD and WER. 

PESQ (Perceptual Evaluation of Speech Quality) \cite{DBLP:conf/icassp/RixBHH01} is an objective speech quality metric originally proposed for telephony quality assessment. It consists of a complex series of filters and transforms which result in a speech quality score. For the purposes of our work, we use this metric to measure how clean a speech signal is.

STOI (Short-Time Objective Intelligibility measure) \cite{DBLP:journals/taslp/TaalHHJ11} aims to measure how intelligible a speech signal is through a comparative DFT-based (Discrete Fourier Transform) approach. It has been found that it achieves close correlation to human intelligibility scores. In our experiments, we use this metric to measure the intelligibility of the reconstructed samples.

MCD (Mel-Cepstral Distance) \cite{407206} is designed to evaluate speech quality based on the cepstrum distance on the mel-scale. In practice, this is calculated as the distance between the MFCCs extracted from two signals. We find that it works quite reliably in measuring perceptual quality in our synthesized outputs, when compared to the original signal. 

WER (Word Error Rate) measures the accuracy of a speech recognition system. It is calculated as:
\begin{align}
    WER = \frac{S+D+I}{N}, 
\end{align}
where $S$ is the number of substitutions, $D$ is the number of deletions, $I$ is the number of insertions and $N$ is the total number of words in an utterance. For our work, we apply pre-trained ASR models to measure WER, which serves as an objective intelligibility metric for the reconstructed speech.

\section{Results on Seen Speakers} \label{section6a}
In this section, we present our experiments for seen speakers. 
For direct comparison with other works we use the same 4 speakers from GRID (1, 2, 4 and 29) as in  \cite{DBLP:conf/icassp/AkbariACM18,DBLP:journals/corr/abs-1906-06301,DBLP:journals/corr/abs-2005-08209,DBLP:journals/corr/abs-2004-02541} and the 3 lipspeakers from TCD-TIMIT as in \cite{DBLP:journals/corr/abs-2005-08209}. In order to investigate the impact of the number of speakers and the amount of training data, we also present results for all 33 speakers from the GRID dataset. We split the utterances in each of these datasets using a 90-5-5\,\% ratio for training, validation and testing respectively similarly to \cite{DBLP:conf/icassp/AkbariACM18,DBLP:journals/corr/abs-1906-06301,DBLP:journals/corr/abs-2005-08209,DBLP:journals/corr/abs-2004-02541}, such that the speakers in the validation and test sets are identical to the speakers seen in the training set (but the utterances are different). To measure the Word Error Rate (WER) for our GRID samples, we use a pre-trained ASR model (based on \cite{ma2019investigating}) which was trained and tested on the full GRID dataset (using the split mentioned in Section \ref{section6b}), achieving a baseline of 4.23\,\% WER on the test set. Audio samples, as well as spectrogram and waveform figures are presented on our website\footnote{\label{note1}\url{https://sites.google.com/view/video-to-speech/home}} for the experiments presented in sections \ref{section6a}, \ref{section6b} and \ref{section6c}. Additionally, we present a publicly available repository\footnote{\url{https://github.com/miraodasilva/evalaudio}} which can be used to reproduce each of the evaluation metrics presented in this work. We are also available to provide generated test samples for researchers hoping to reproduce or compare with our work.

\subsection{Ablation Study}
Results for the ablation study are shown in Table \ref{ablation_seen}. For this study, we only consider the 4 subjects from GRID presented above (1,2,4 and 29).

 Firstly, we observe that each of the three comparative losses $L_{PASE}$, $L_{power}$ and $L_{MFCC}$ yield considerable improvements in the verbal accuracy of samples (as shown by the WER), even when only one is removed. We can also observe that $L_{MFCC}$ and $L_{power}$ are particularly impactful on the MCD of the reported samples, which is unsurprising since this is an MFCC-based metric. On the other hand, it is clear that $L_{PASE}$ is essential towards achieving high intelligibility, given its particular impact on STOI. Finally, all three losses also seem to positively impact the PESQ score, indicating an increase in overall audio clarity. 
 
 We can see that the simultaneous removal of $L_{PASE}$ and $L_{power}$ greatly decreases PESQ and STOI, indicating that these losses are particularly important towards the clarity of generated samples. We also show that the absence of $L_{MFCC}$ and $L_{power}$ sharply increases MCD, indicating that these two losses greatly increase the similarity between real and synthesized audio. On the other hand, this model maintains a WER below 10\,\%, which means that $L_{PASE}$ alone (together with the adversarial losses) can achieve intelligible audio. Finally, the removal of all three L1 losses results in realistic yet unintelligible audio. This is because the adversarial losses are the only objective used for training, and therefore there is no incentive for the network to learn the exact words corresponding to the input video.

 We observe that the use of the waveform critic yields noticeable improvements through our metrics, particularly in WER and STOI, suggesting that its inclusion substantially increases intelligibility. Additionally, the power critic also yields moderate improvements in PESQ, STOI and WER. Finally, we observe that the removal of both critics results in substantially lower MCD and WER, but mantains PESQ and STOI at a similar value. This again indicates that our model can generate intelligible and accurate words without the adversarial losses. However, these synthesized samples lack realism, which drastically improves when the critics are used. To demonstrate this effect, readers are encouraged to listen to examples on our website$^{\ref{note1}}$.
 
We also experiment with using the full face as input, as this is commonly used in previous studies. Through this ablation, we show that using a cropped mouth region instead of the full face improves our results substantially regarding WER, effectively improving intelligibility. We also prove that the use of overlap improves all metrics slightly, suggesting that its purpose of minimizing the issue of frame transitions is benefitial towards output quality. 

\begin{figure*}
\centering
     \begin{subfigure}{0.3\linewidth}
        \includegraphics[width=\linewidth]{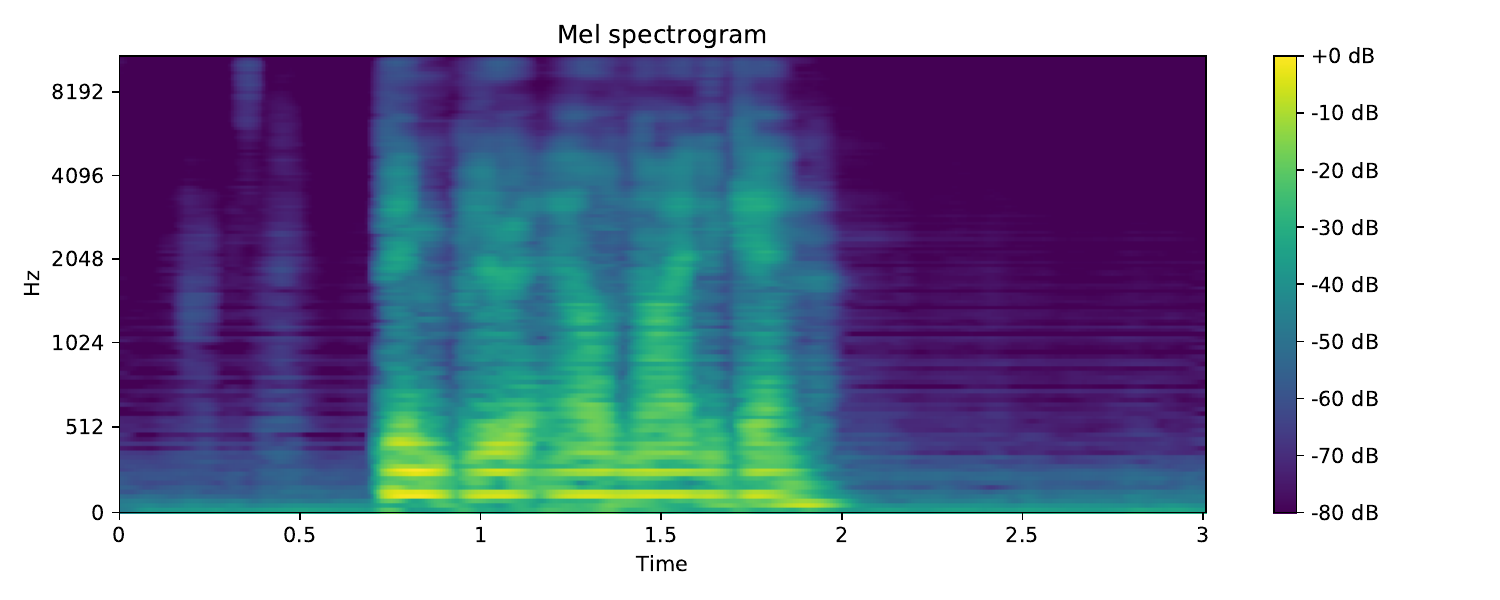}
        \caption{w/o Wave Critic, w/o Power Critic}
    \end{subfigure}
    \begin{subfigure}{0.3\linewidth}
        \includegraphics[width=\linewidth]{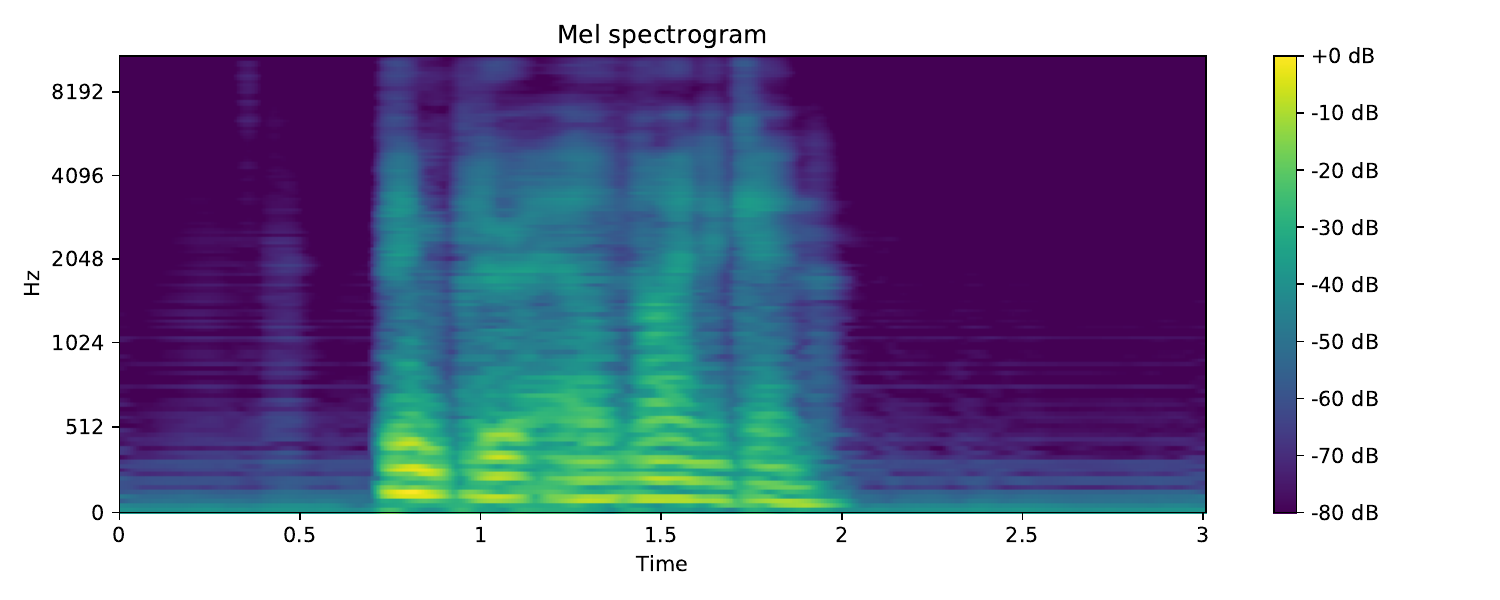}
        \caption{Full Model}
    \end{subfigure}
    \begin{subfigure}{0.3\linewidth}
        \includegraphics[width=\linewidth]{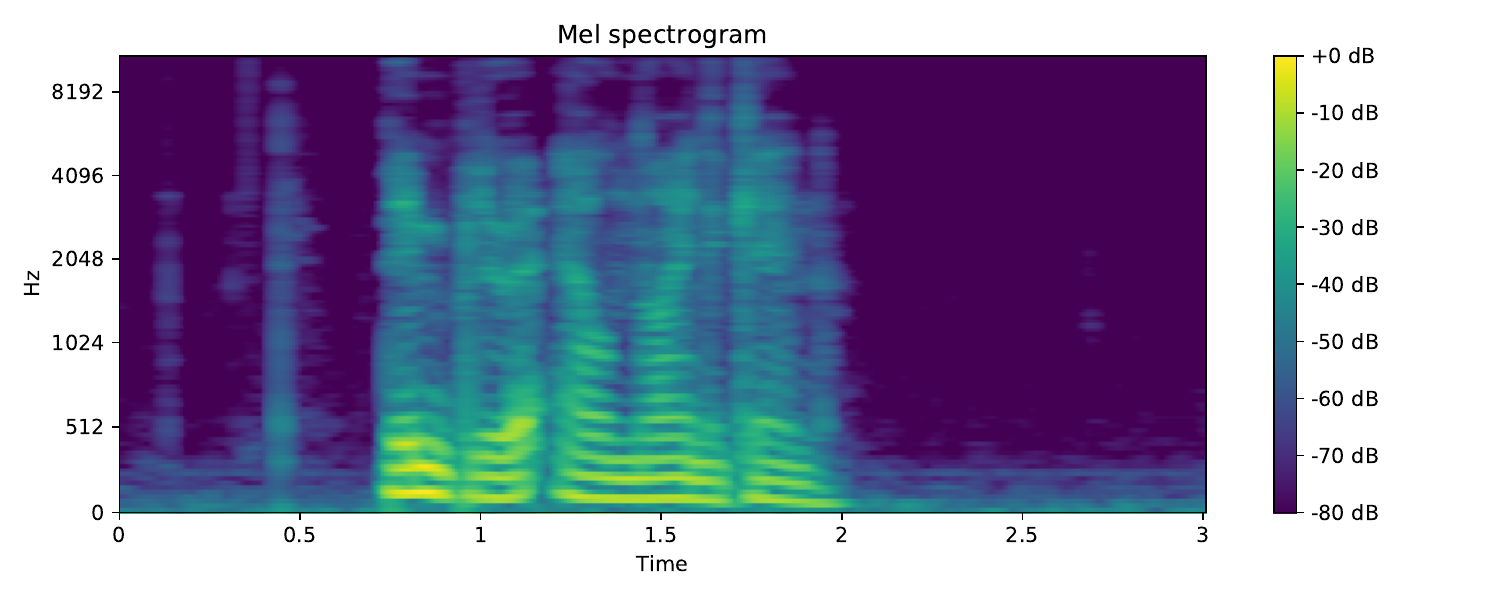}
        \caption{Real Audio}
    \end{subfigure}
\caption{Mel-frequency spectrograms taken from the audio reconstructed with our seen speaker ablation models. The clip we present is from GRID, speaker 1, utterance 'Bin blue at L 9 again'.}
\label{ablation_qualitative}
\end{figure*}

A qualitative comparison with other works can be seen in Figure \ref{ablation_qualitative}. Compared to the real audio, our spectrogram is similar overall, but is slightly blurrier and fails to model some of the fine details in the frequency bins, especially in the higher frequencies. The model trained without adversarial critics features a much blurrier spectrogram than the full model, failing to reproduce even the lower frequency bands during voiced speech, highlighting the importance of adversarial training.

\begin{table}
\centering 
\def\arraystretch{1.5}
\begin{tabular}{  m{3.5cm} | >{\centering}m{0.8cm}  >{\centering}m{0.8cm}  >{\centering}m{0.8cm}  >{\centering\arraybackslash}m{0.8cm} } 
\hline
Model & PESQ & STOI & MCD & WER  \\
\hline
\hline
w/o $L_{PASE}$ & 2.06 & \textbf{0.597} & \textbf{26.44} & 8.97\,\% \\
\hline
w/o $L_{power}$ & 2.05 & 0.575 & 28.64 & 9.54\,\% \\
\hline
w/o $L_{MFCC}$ & 2.08 & 0.591 & 28.09 & 9.09\,\% \\
\hline
w/o $L_{PASE}$, w/o $L_{power}$ & 1.86 & 0.545 & 27.47 & 13.44\,\%  \\
\hline
w/o $L_{PASE}$, w/o $L_{MFCC}$ & 2.02 & 0.589 & 28.82 & 13.33\,\%  \\
\hline
w/o $L_{MFCC}$, w/o $L_{power}$ & 2.00 & 0.569 & 31.43 & 9.71\,\%  \\
\hline
w/o $L_{PASE}$, w/o $L_{power}$, w/o $L_{MFCC}$ & 1.14 & 0.311 & 53.63 & 89.12\,\%  \\
\hline
w/o waveform critic & 2.07 & 0.583 & 26.66 & 8.47\,\% \\
\hline
w/o power critic & 2.08 & 0.594 & 26.73 & 7.30\,\% \\
\hline
w/o waveform critic, w/o power critic & 2.07 & 0.584 & 27.45 & 9.01\,\% \\
\hline
w/o overlap & 2.06 & 0.590 & 26.73 & 7.40\,\% \\
\hline
w/ full face & 2.07 & 0.596 & 26.46 & 9.94\,\% \\
\hline
full model & \textbf{2.10} & 0.595 & 26.78 & \textbf{7.03}\,\% \\
\hline
\end{tabular}
\caption{Ablation study performed on GRID for seen speaker speech reconstruction.}
\label{ablation_seen}
\end{table}

\subsection{Comparison with Other Works}
We compare our proposed model with previous works on the commonly used 4 GRID speakers as shown in Table \ref{grid_seen_comparison}. We note that the metrics reported on Lip2Wav \cite{DBLP:journals/corr/abs-2005-08209} are taken directly from their paper due to test samples not being publicly available, and that their WER was calculated using the Google Speech-to-Text (STT) API rather than our ASR model.

Regarding PESQ, it is clear that our model is superior to the previous approaches by a sizeable margin. This suggests that the quality of our synthesized speech is somewhat higher than past models. Our model also outperforms previous works on STOI, excluding Lip2Wav. This shows that our samples are more intelligible than most other approaches, but are outperformed by the robustness and consistency of the speech produced by Lip2Wav. Furthermore, our generated samples achieve a better MCD than previous works, indicating that our reconstructed audio is more accurate than previous approaches on the frequency domain. Finally, our work achieves the best WER out of all methods, which shows that our model is more accurate than any of the previous approaches by a large factor, outperforming our previous model by more than 10\,\%.

\begin{figure*}
\centering
    \begin{subfigure}{\linewidth}
        \centering
        \begin{minipage}[c]{0.01\linewidth}
            \centering
            \caption{}
        \end{minipage}
        \begin{minipage}[c]{0.95\linewidth}
            \centering
            \begin{subfigure}{0.3\linewidth}
                \includegraphics[width=\linewidth]{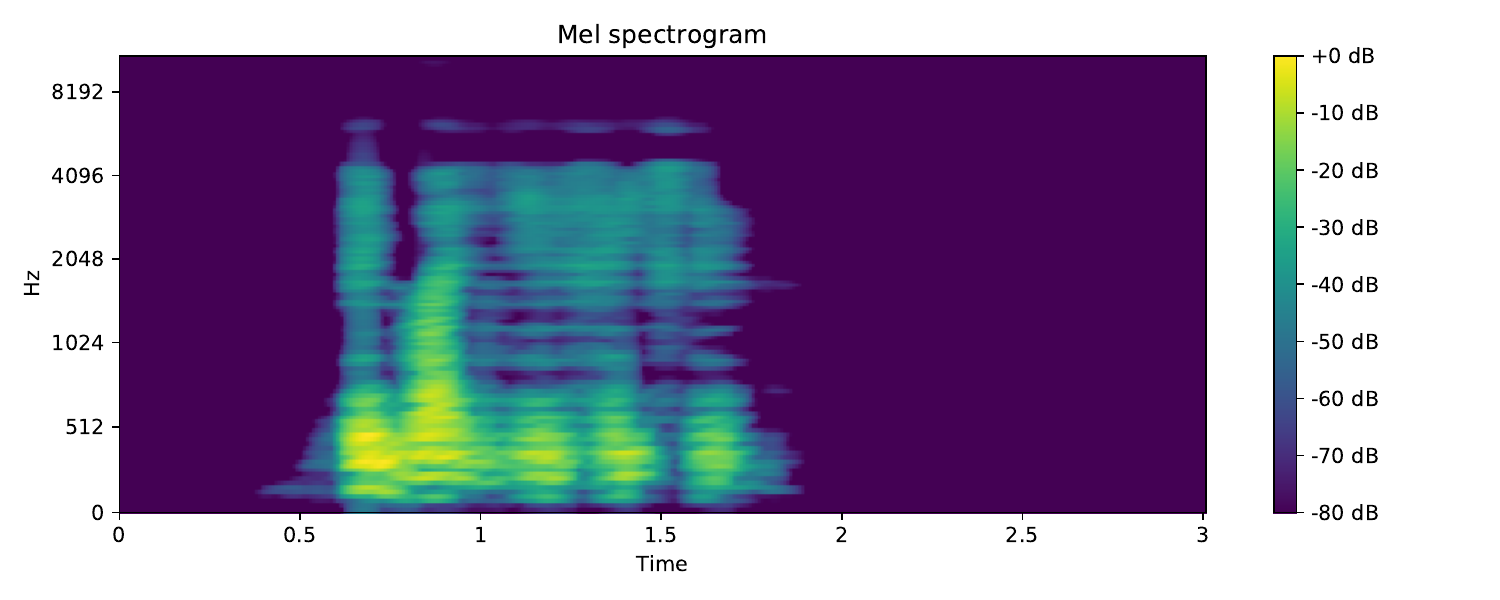}
            \end{subfigure}
            \begin{subfigure}{0.3\linewidth}
                \includegraphics[width=\linewidth]{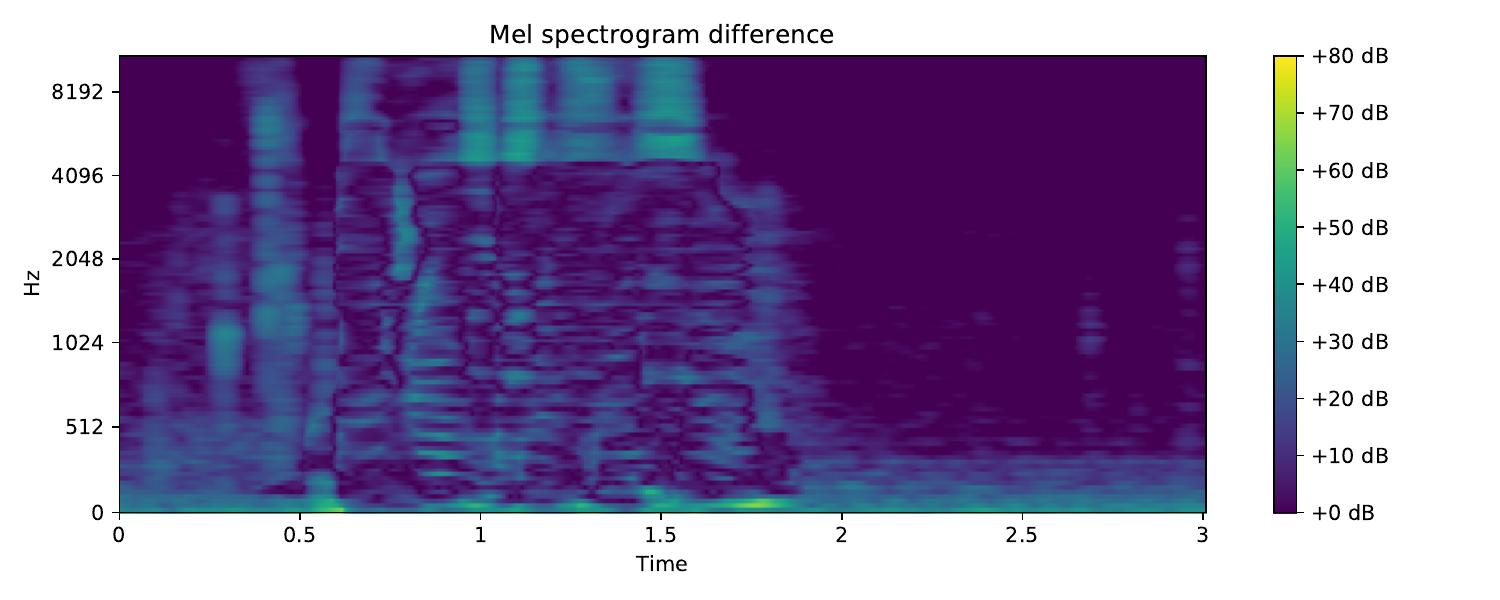}
            \end{subfigure}
            \begin{subfigure}{0.3\linewidth}
                \includegraphics[width=\linewidth]{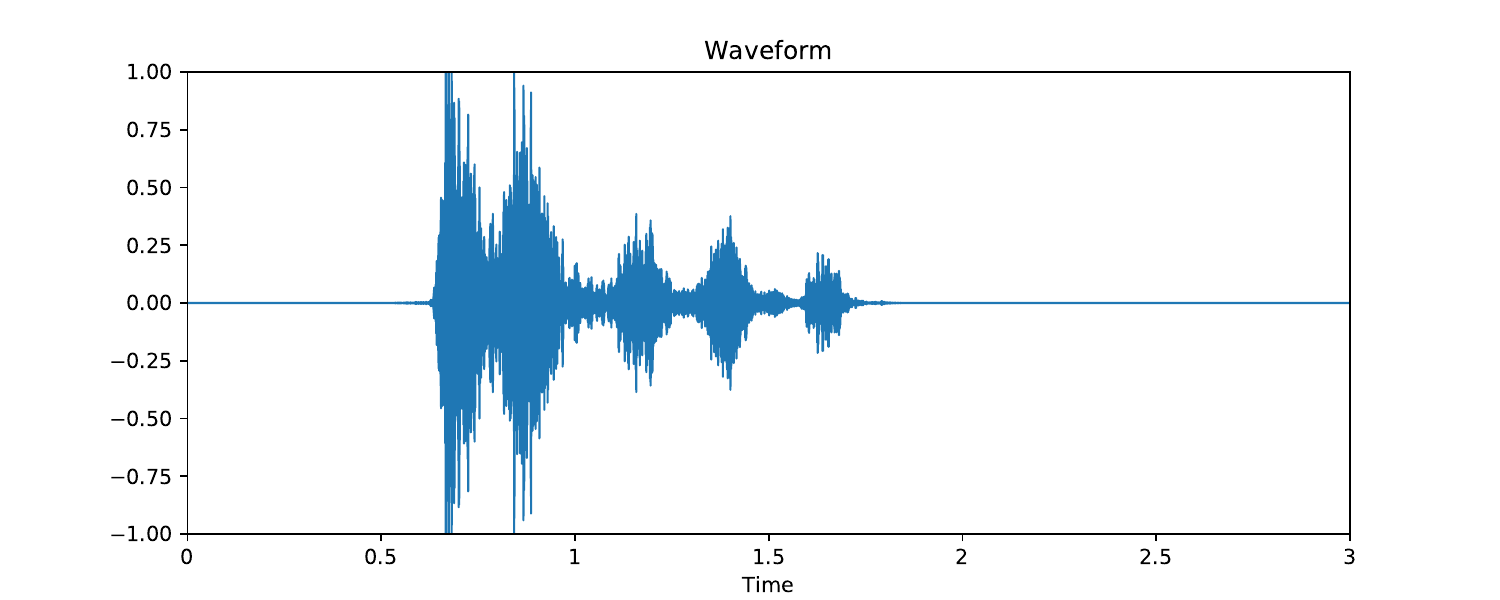}
            \end{subfigure}
        \end{minipage}
    \end{subfigure}
    \begin{subfigure}{\linewidth}
        \centering
        \begin{minipage}[c]{0.01\linewidth}
            \centering
            \caption{}
        \end{minipage}
        \begin{minipage}[c]{0.95\linewidth}
            \centering
            \begin{subfigure}{0.3\linewidth}
                \includegraphics[width=\linewidth]{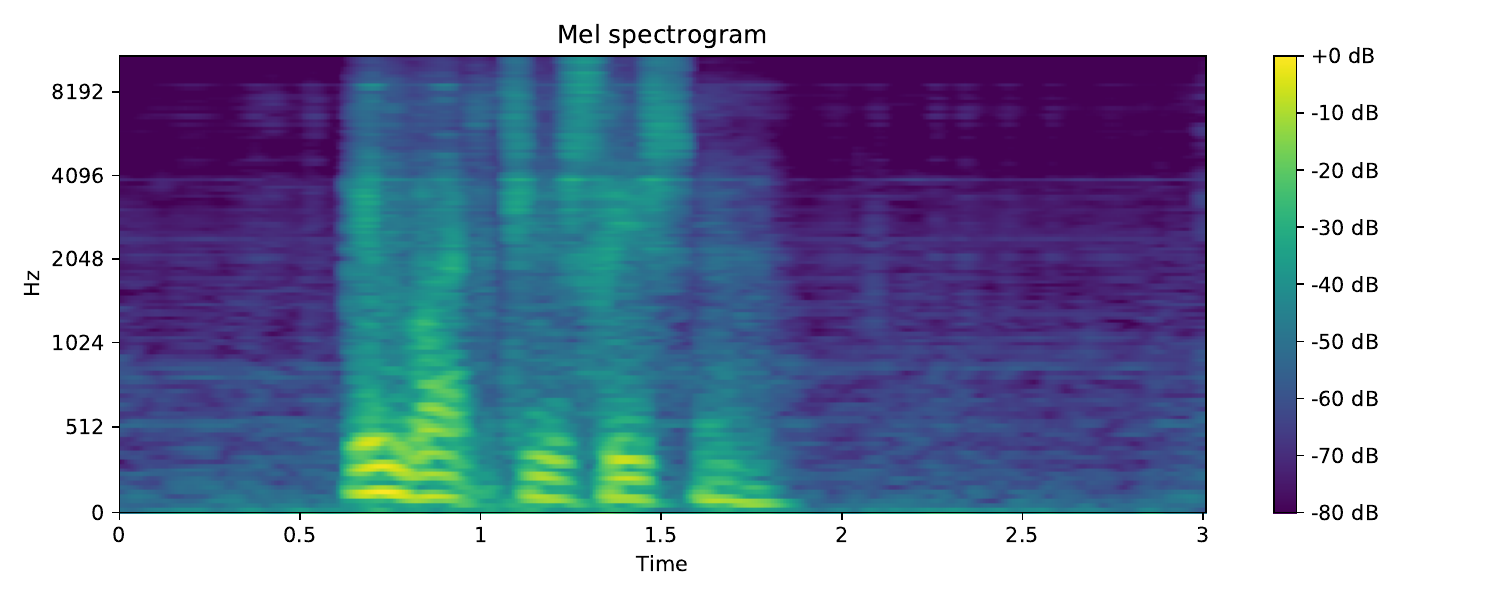}
            \end{subfigure}
            \begin{subfigure}{0.3\linewidth}
                \includegraphics[width=\linewidth]{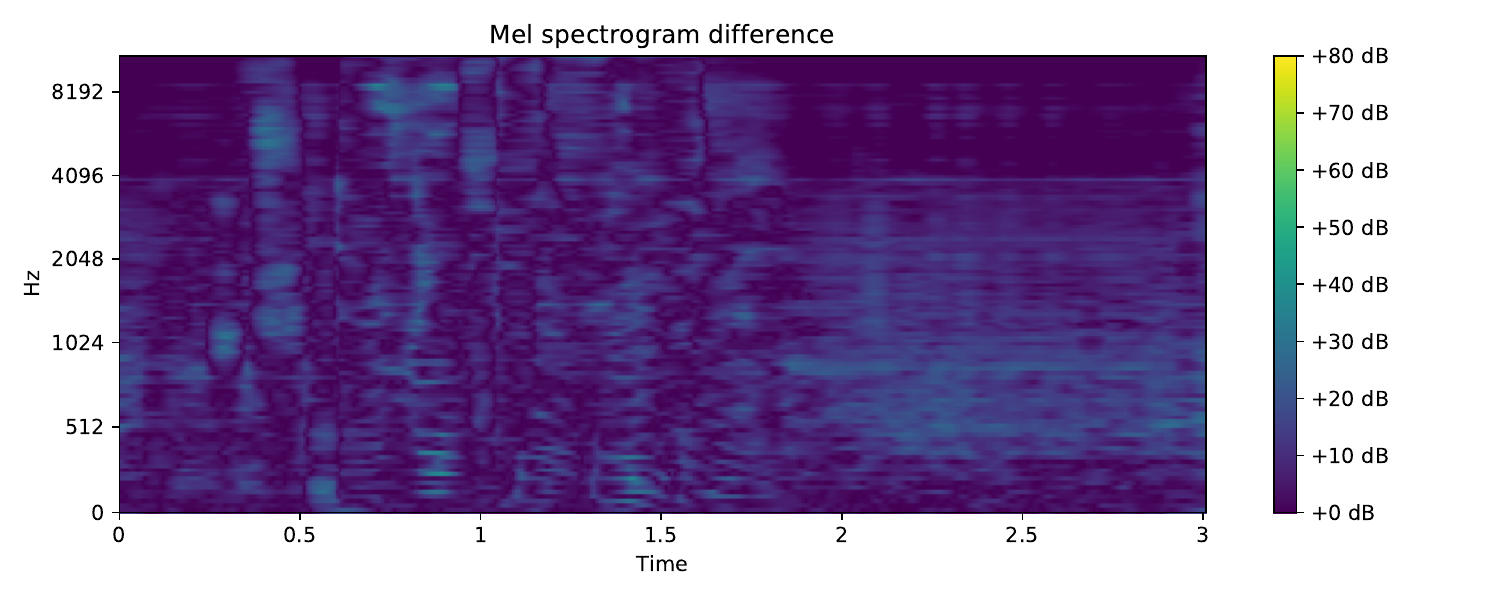}
            \end{subfigure}
            \begin{subfigure}{0.3\linewidth}
                \includegraphics[width=\linewidth]{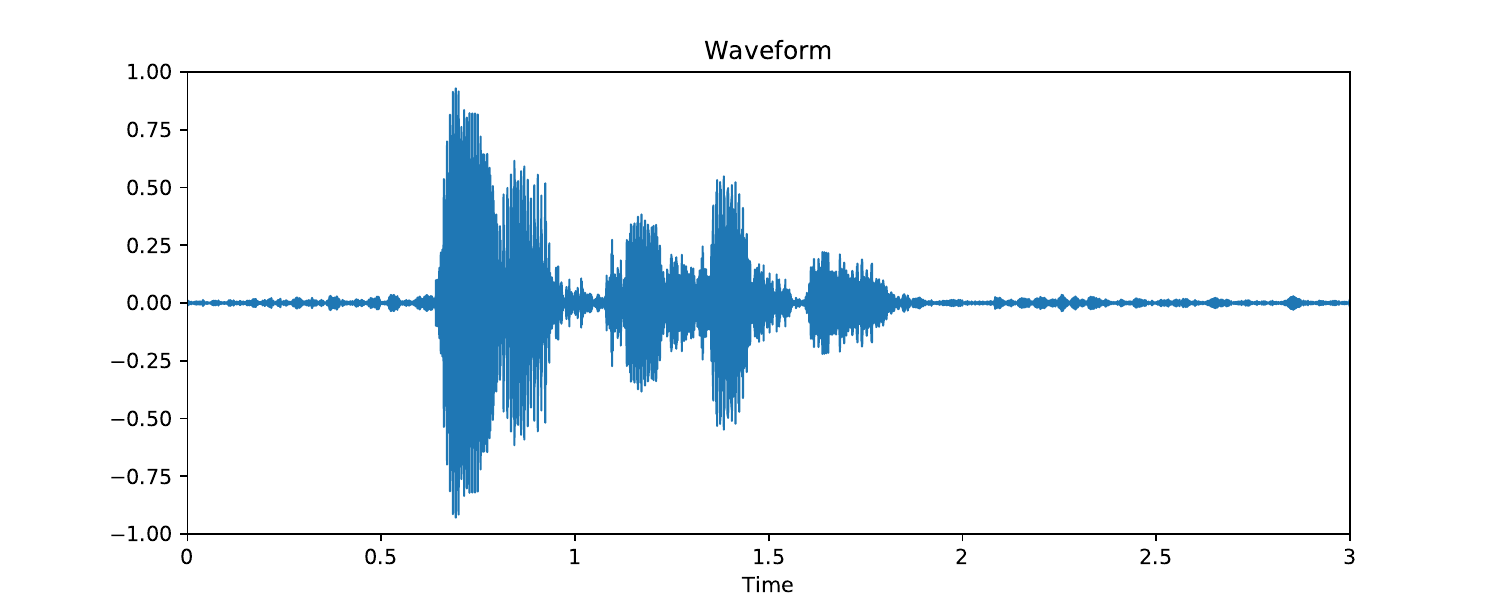}
            \end{subfigure}
        \end{minipage}
    \end{subfigure}
    \begin{subfigure}{\linewidth}
        \centering
        \begin{minipage}[c]{0.01\linewidth}
            \centering
            \caption{}
        \end{minipage}
        \begin{minipage}[c]{0.95\linewidth}
            \centering
            \begin{subfigure}{0.3\linewidth}
                \includegraphics[width=\linewidth]{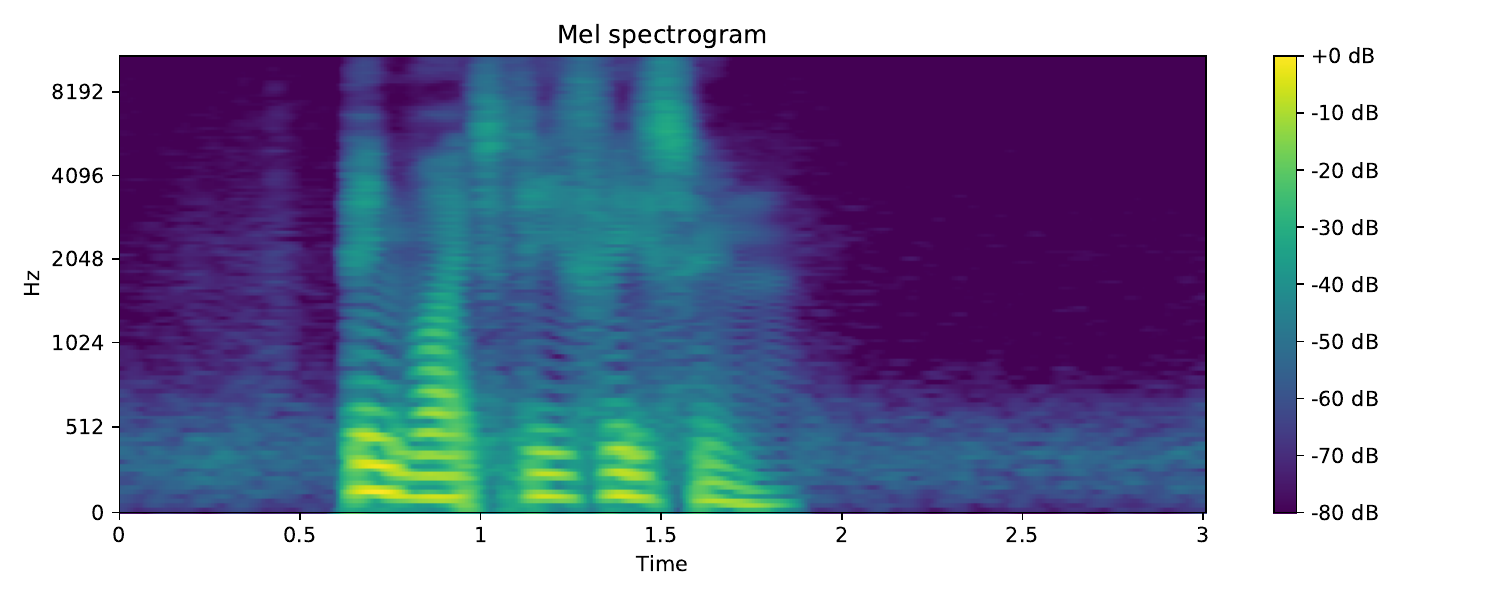}
            \end{subfigure}
            \begin{subfigure}{0.3\linewidth}
                \includegraphics[width=\linewidth]{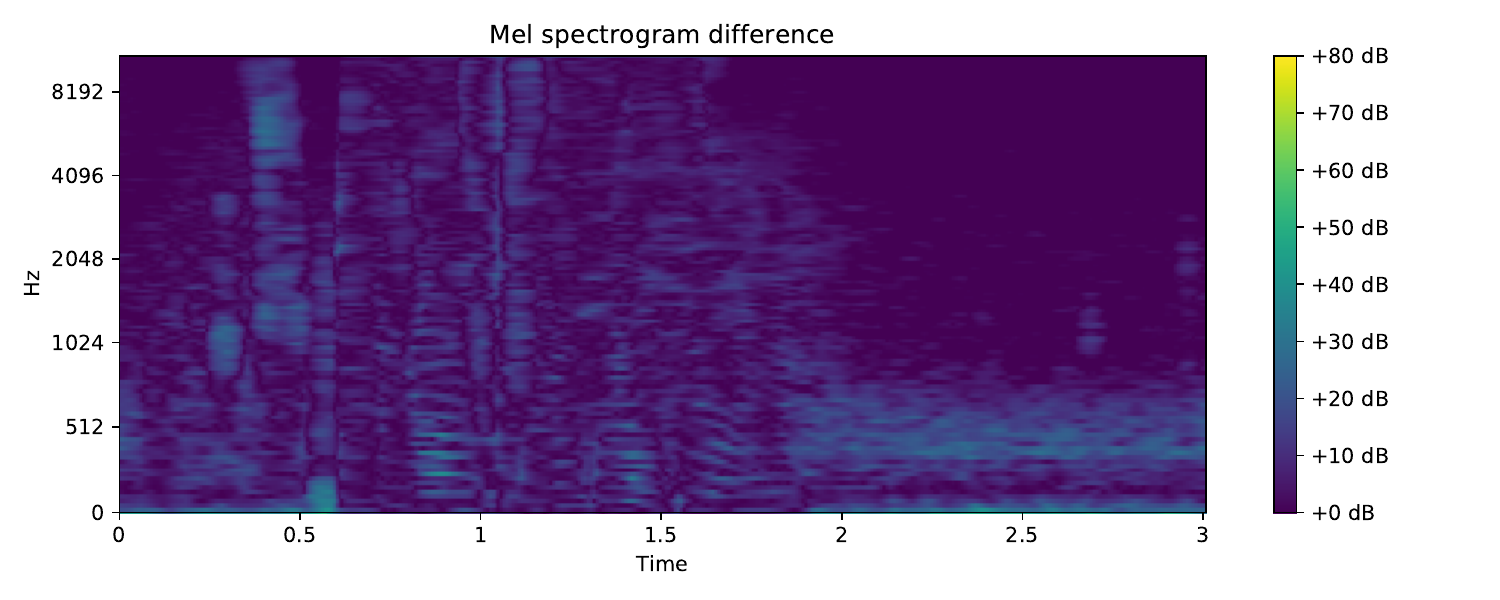}
            \end{subfigure}
            \begin{subfigure}{0.3\linewidth}
                \includegraphics[width=\linewidth]{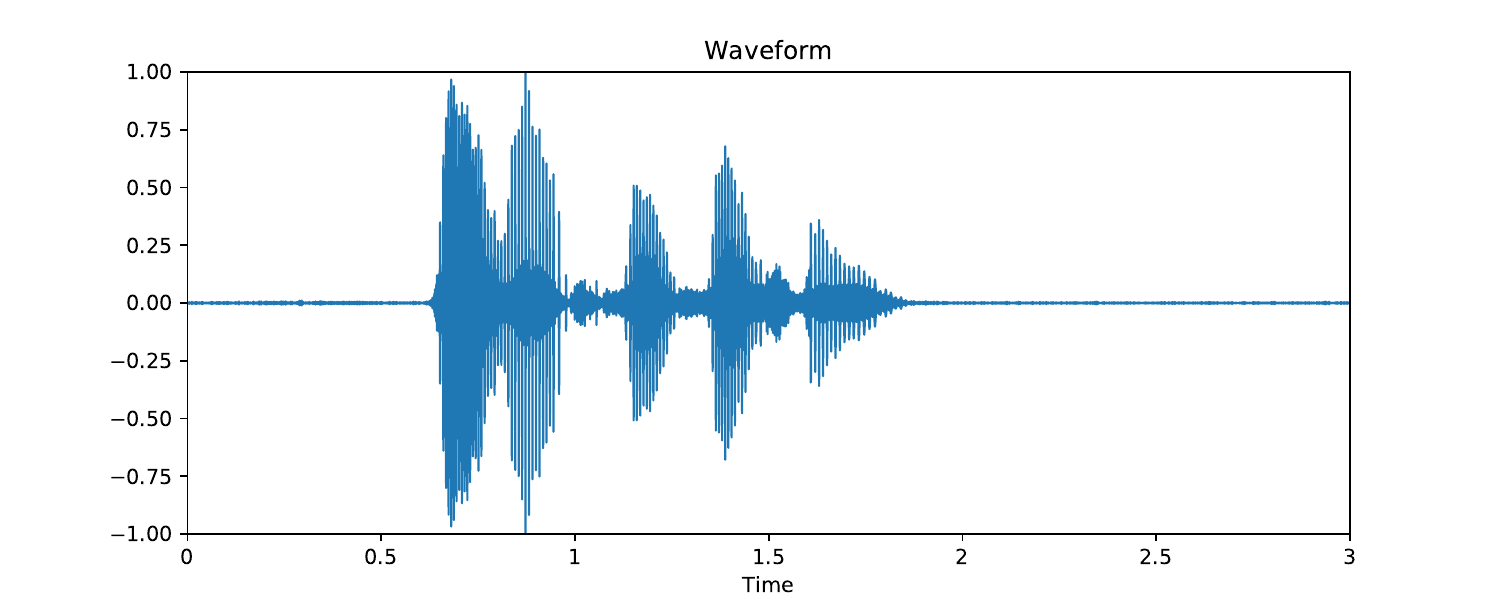}
            \end{subfigure}
        \end{minipage}
    \end{subfigure}
    \begin{subfigure}{\linewidth}
        \centering
        \begin{minipage}[c]{0.01\linewidth}
            \centering
            \caption{}
        \end{minipage}
        \begin{minipage}[c]{0.95\linewidth}
            \centering
            \begin{subfigure}{0.3\linewidth}
                \includegraphics[width=\linewidth]{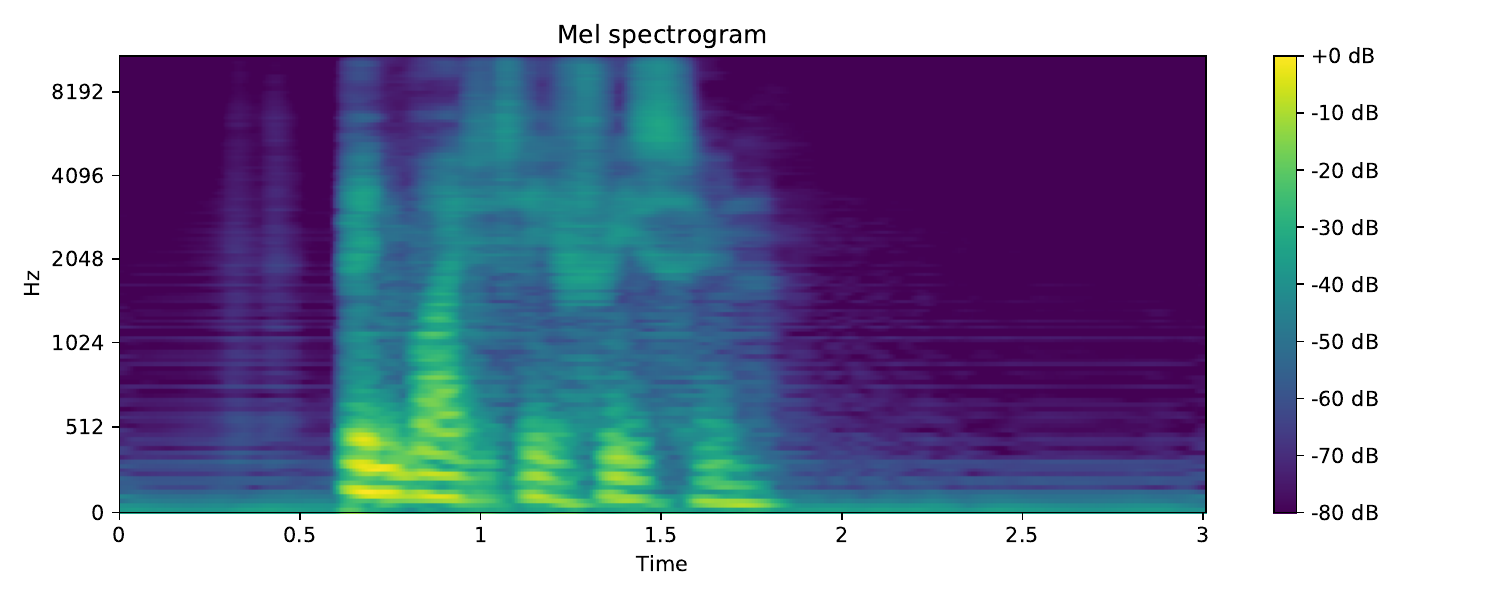}
            \end{subfigure}
            \begin{subfigure}{0.3\linewidth}
                \includegraphics[width=\linewidth]{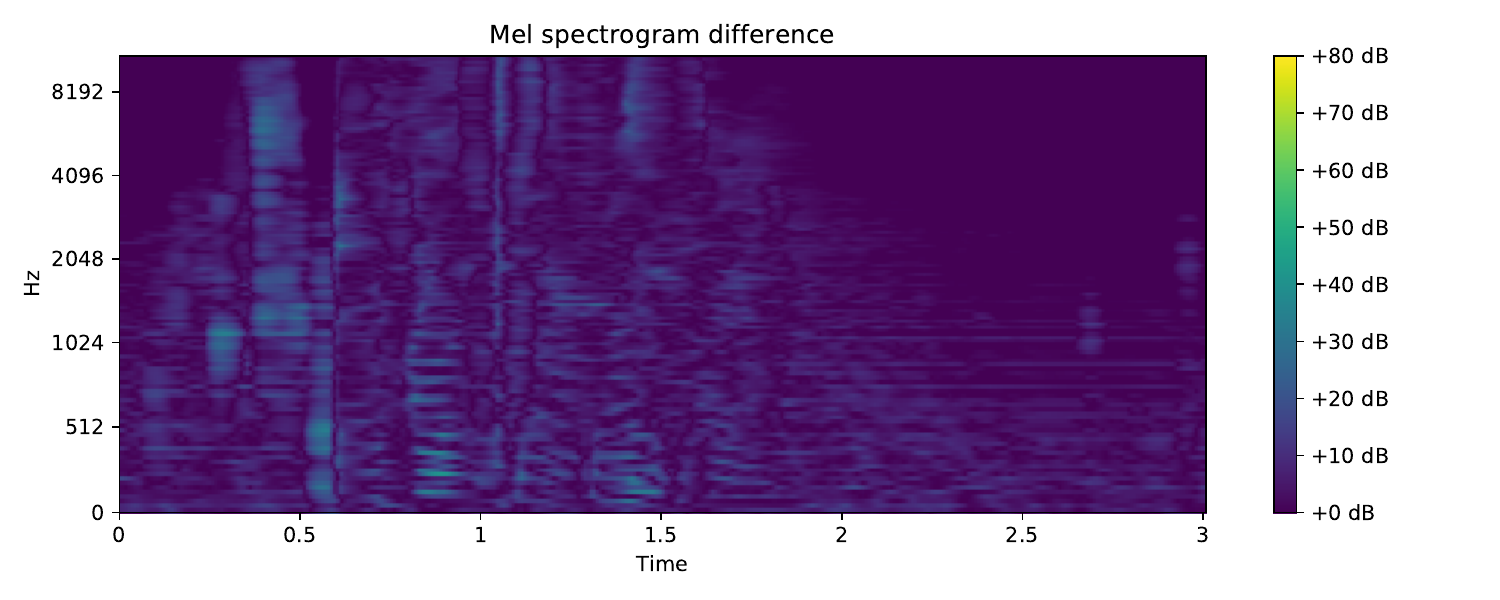}
            \end{subfigure}
            \begin{subfigure}{0.3\linewidth}
                \includegraphics[width=\linewidth]{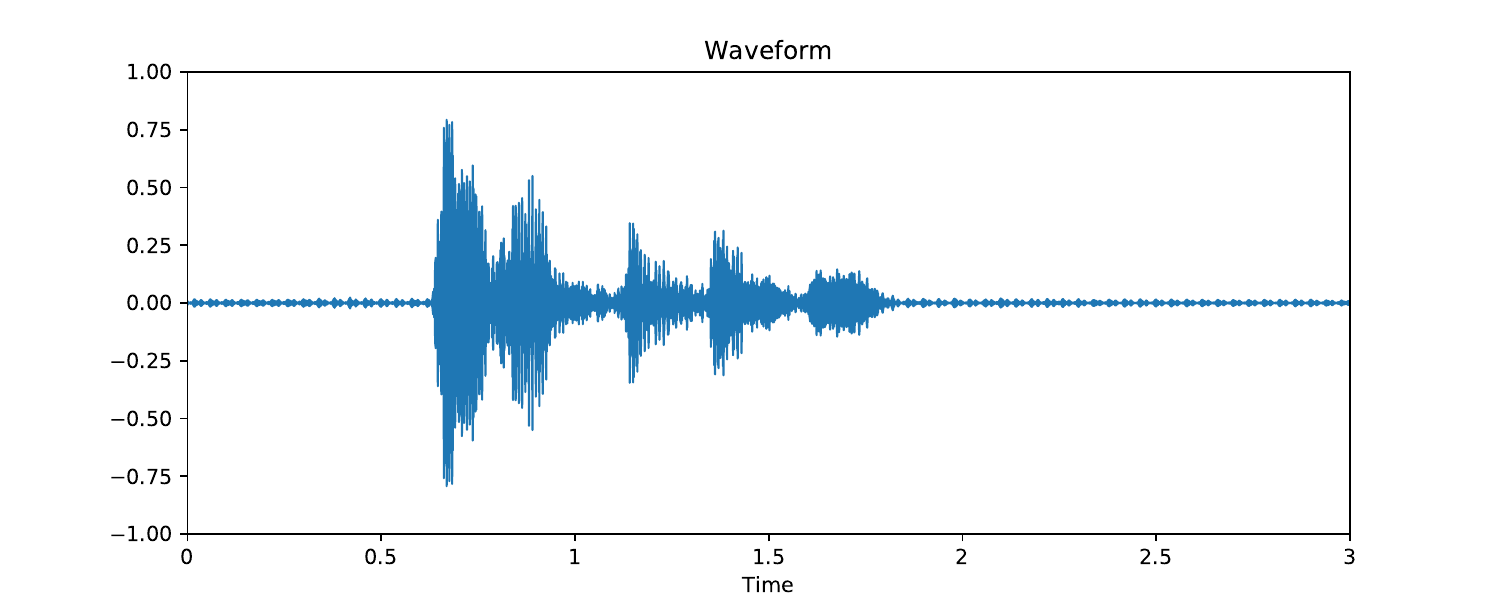}
            \end{subfigure}
        \end{minipage}
    \end{subfigure}
    \begin{subfigure}{\linewidth}
        \hspace{0.5em}
        \begin{minipage}[c]{0.01\linewidth}
            \centering
            \caption{}
        \end{minipage}
        \begin{minipage}[c]{0.95\linewidth}
            \centering
            \begin{subfigure}{0.3\linewidth}
                \includegraphics[width=\linewidth]{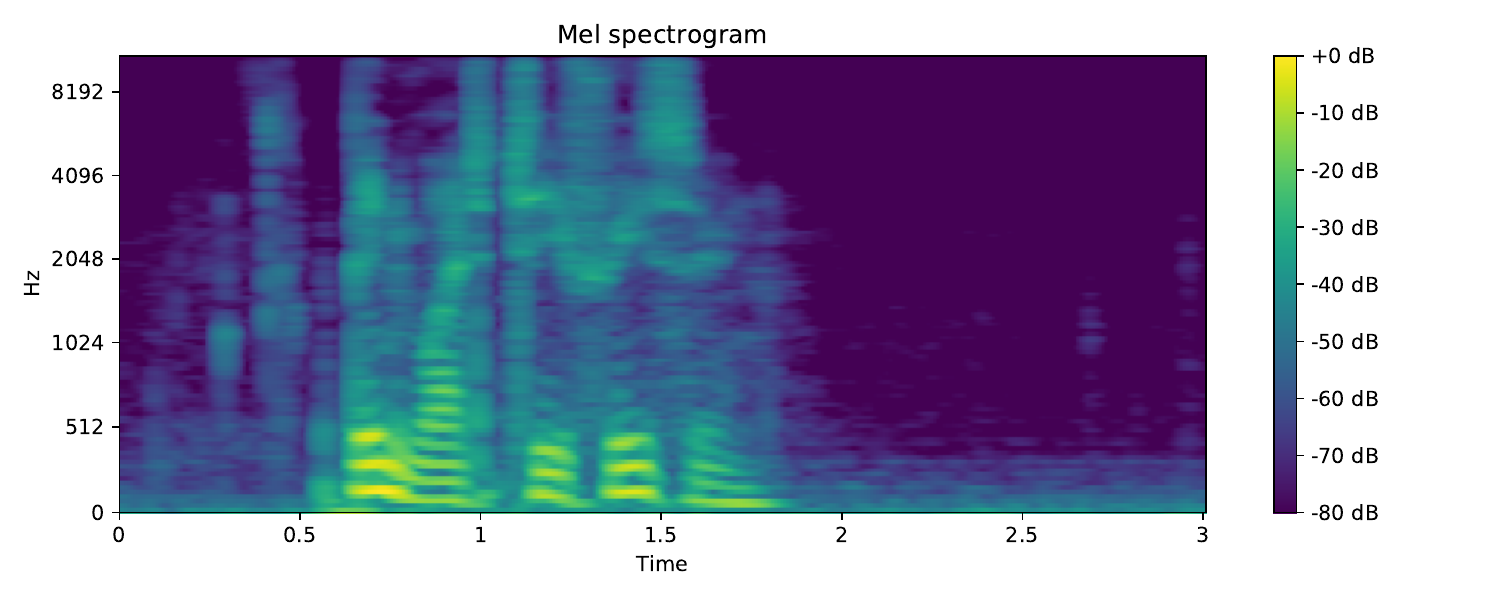}
            \end{subfigure}
            \begin{subfigure}{0.3\linewidth}
                \includegraphics[width=\linewidth]{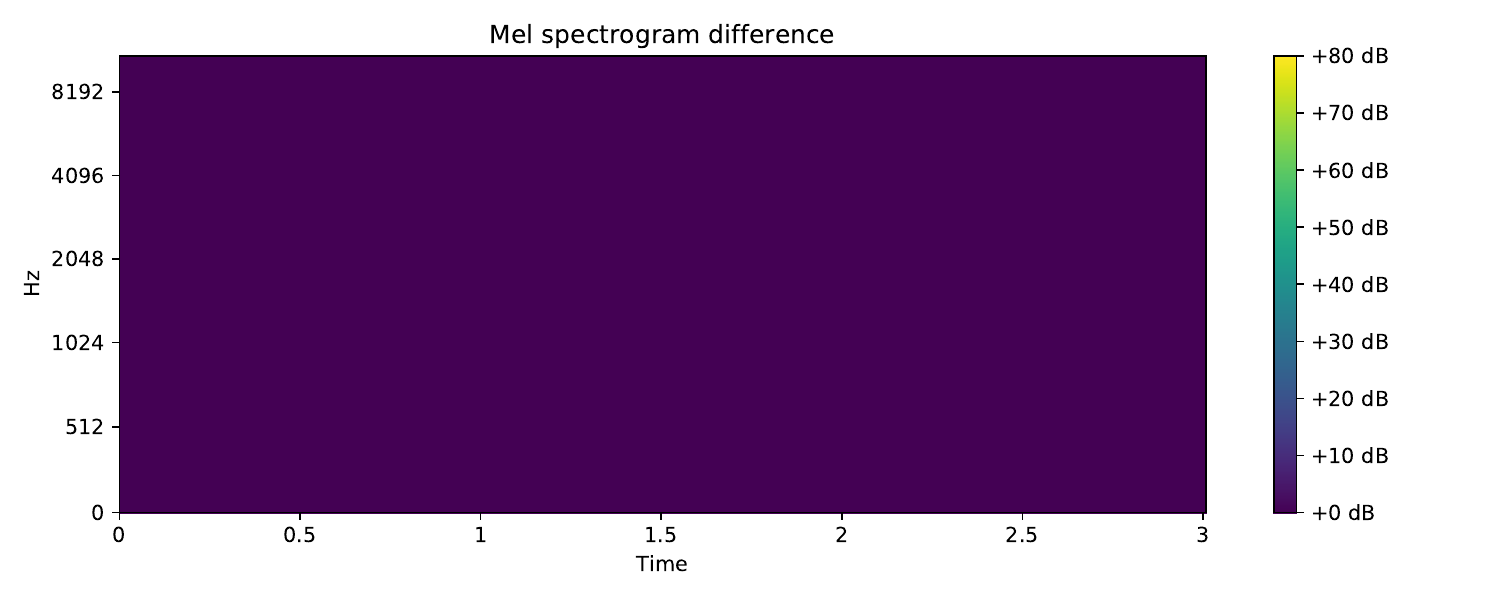}
            \end{subfigure}
            \begin{subfigure}{0.3\linewidth}
                \includegraphics[width=\linewidth]{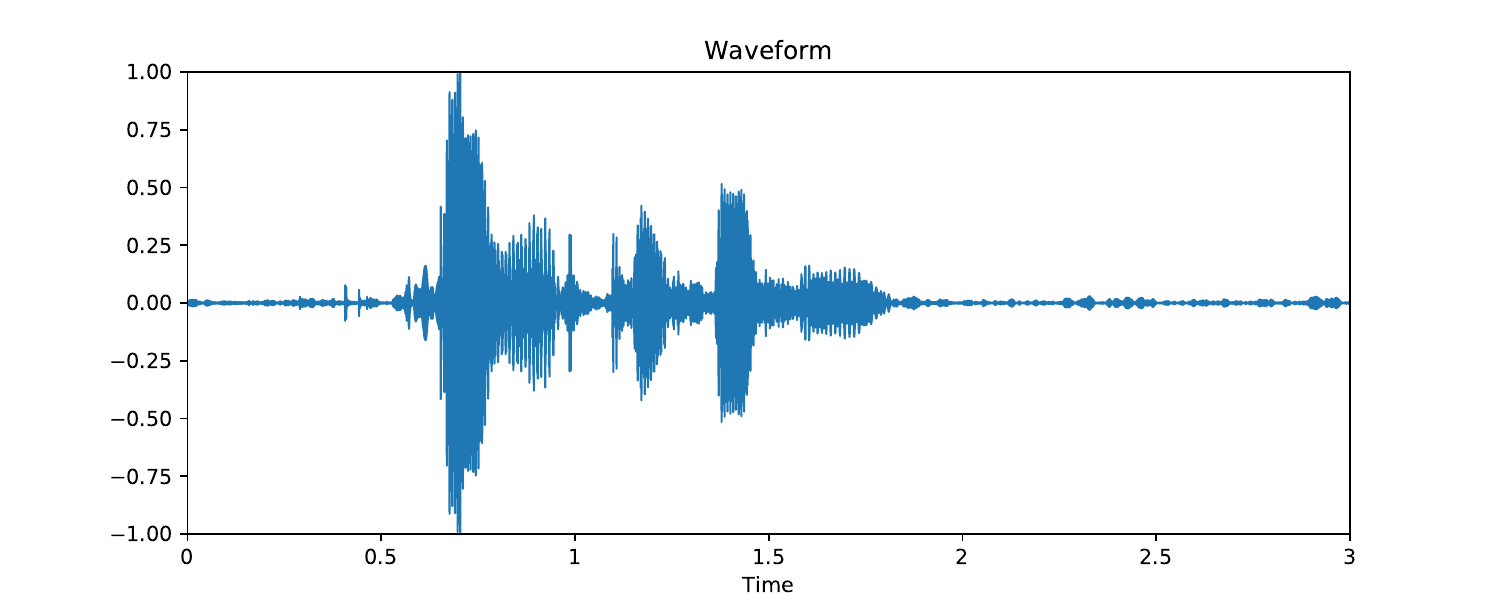}
            \end{subfigure}
         \end{minipage}
    \end{subfigure}
\caption{Mel-frequency spectrograms (left), Mel-frequency spectrogram differences (middle) and waveforms (right) taken from the audio reconstructed with Lip2AudSpec \cite{DBLP:conf/icassp/AkbariACM18} (a), our previous work \cite{DBLP:journals/corr/abs-1906-06301} (b), a previous vocoder-based model \cite{DBLP:journals/corr/abs-2004-02541} (c) and our model (d), as well as the real audio (e) -- GRID, Speaker 1, utterance 'Bin white at T 3 soon'. All models were trained on the same split of GRID (4 speakers, seen speaker split), as presented in our comparison.}
\label{qualitative_comparison}
\end{figure*}

A qualitative comparison is shown in Figure \ref{qualitative_comparison}, which displays waveforms, mel-frequency spectrogram, and mel-frequency spectrogram differences, \ie the element-wise absolute difference between the real and synthesized spectrograms. This difference is calculated as:
\begin{align}
    \| MelSpec(x) - MelSpec(\widetilde{x}) \|,
\end{align}
where $x$ is the real waveform and $\widetilde{x}$ is the synthesized waveform. Through the spectrograms, it is clear that Lip2Audspec is the least accurate in the frequency domain, failing to model many frequencies, particularly in the higher bands. The other three approaches are clearly more accurate, but all feature some inaccuracies during voiced speech and also noise in unvoiced segments. While \cite{DBLP:journals/corr/abs-1906-06301} and \cite{DBLP:journals/corr/abs-2004-02541} feature an excessive amount of low frequency noise, our model seems to accurately emulate the low amount of noise in the real audio and therefore achieves the least substantial spectrogram difference.

We also compare our model to Lip2Wav on TCD-TIMIT (3 lipspeakers) in Table \ref{timit_comparison}. Once more, it is clear that our model outperforms Lip2Wav \cite{DBLP:journals/corr/abs-2005-08209} on PESQ, but achieves lower performance on STOI, which indicates that our model produces clearer, yet somewhat less intelligible audio. Additionally, our samples achieve a reasonably low MCD, indicating moderate similarity in the frequency domain.

\begin{table}
\centering 
\def\arraystretch{1.5}
\begin{minipage}{\linewidth}
\begin{tabular}{  m{2.5cm} | >{\centering}m{1cm}  >{\centering}m{1cm}  >{\centering}m{1cm}  >{\centering\arraybackslash}m{1.1cm} } 
\hline
Method & PESQ & STOI & MCD & WER \\
\hline
\hline
Lip2Audspec \cite{DBLP:conf/icassp/AkbariACM18} & 1.81 & 0.425 & 63.88 & 46.36\,\%  \\
\hline
GAN-based \cite{DBLP:journals/corr/abs-1906-06301} & 1.70 & 0.539 & 45.37 & 21.11\,\%  \\
\hline
Vocoder-based \cite{DBLP:journals/corr/abs-2004-02541} & 1.90 & 0.553 & 46.64 & 22.14\,\%  \\
\hline
Lip2Wav \cite{DBLP:journals/corr/abs-2005-08209} & 1.77 & \textbf{0.731} & - & 14.08\footnote{Reported using Google STT API.}\,\% \\  
\hline
Ours & \textbf{2.10} & 0.595 & \textbf{26.78} & \textbf{7.03\,\%}  \\
\hline
\end{tabular}
\end{minipage}
\caption{Comparison between our model and previous works, using the GRID subset (4 speakers) with a seen speaker split. }
\label{grid_seen_comparison}
\end{table}
\begin{table}
\centering 
\def\arraystretch{1.5}
\begin{tabular}{  m{2.5cm} | >{\centering}m{1cm}  >{\centering}m{1cm}  >{\centering\arraybackslash}m{1.1cm} } 
\hline
Method & PESQ & STOI & MCD \\
\hline
\hline
Lip2Wav \cite{DBLP:journals/corr/abs-2005-08209} &  1.35 & \textbf{0.558} & - \\
\hline
Ours & \textbf{1.61} & 0.295 & \textbf{32.12} \\
\hline
\end{tabular}
\caption{Comparison between our model and Lip2Wav, using TCD-Timit (3 lipspeakers) with a seen speaker split.}
\label{timit_comparison}
\end{table}

\subsection{Performance as a Function of Training Set Size}
For the purposes of this study, we use all 33 subjects from GRID and we report results as we vary the size of the training set from 20\,\% to 100\,\% in steps of 20\,\%. Results are shown in Table \ref{varying_training_set}. When compared to the results reported for GRID (4 speakers, seen split), we observe comparable performance for 33 speakers when using the full training set. This shows that our network adapts well to larger datasets and is able to model a large amount of speakers with no substantial drop in performance.

Regarding the models which are trained using a smaller subset of the training set, it is clear that the performance drops as the amount of training data is gradually reduced. However, it is worth highlighting that the overall performance remains moderately consistent, even when we use only 20\,\% of the training data. This shows that our model adapts well to smaller datasets. We note that all 5 models were trained for the same amount of total training steps to avoid any bias in our comparative results.
\begin{table}
\centering 
\def\arraystretch{1.5}
\begin{tabular}{  m{2cm} | >{\centering}m{1cm}  >{\centering}m{1cm}  >{\centering}m{1cm}  >{\centering\arraybackslash}m{1.1cm} } 
\hline
\% of Training Set & PESQ & STOI & MCD & WER  \\
\hline
\hline
20\,\% & 1.96 & 0.583 & 29.22 & 11.78\,\%  \\
\hline
40\,\% & 2.00 & 0.594 & 28.49 & 10.10\,\%  \\
\hline
60\,\% & 2.02 & 0.595 & 27.94 & 9.06\,\%  \\
\hline
80\,\% & 2.02 & 0.596 & 27.68 & 8.36\,\%  \\
\hline
100\,\% & 2.02 & 0.601 & 27.78 & 8.03\,\%  \\
\hline
\end{tabular}
\caption{Study on the performance of our speech reconstruction model using varying training set sizes, using the full GRID seen speaker split mentioned in Section \ref{section4}.}
\label{varying_training_set}
\end{table}

\section{Results on Unseen Speakers} \label{section6b}
In this section, we investigate the performance of the proposed approach on unseen speakers. For the purposes of this study, we use all speakers from the GRID dataset, using a 50-20-30\,\% split ratio similarly to \cite{DBLP:journals/corr/abs-1906-06301,DBLP:journals/corr/abs-2004-02541}, such that there is no overlap between the speakers featured in the training, validation and test sets.
To measure WER, we use the GRID pre-trained model mentioned in the previous section.

\subsection{Ablation Study}
In this study, we use all 33 GRID speakers. The results for the ablation study are shown in Table \ref{ablation_unseen}.

For this, task, we find that $L_{power}$ provides the greatest impact on the quality of results, providing a substantial improvement in all metrics. On the other hand, $L_{PASE}$ and $L_{MFCC}$ show noticeable improvements in PESQ and STOI, indicating that these losses contribute to the clarity and intelligibility of the generated samples.  Furthermore, we once more find that $L_{MFCC}$ and $L_{power}$ are particularly important towards achieving a low MCD, meaning that these losses are essential towards achieving accurate MFCCs in our synthesized samples. 

Regarding the adversarial loss, we can see that, as reported in the seen speaker ablation, PESQ, STOI and MCD improve with the addition of the waveform and power critics. This suggests that these critics have a positive effect on the clarity and intelligiblity of samples, and that the accuracy on the frequency domain is improved as well. However, we observe that the WER remains at a similar value with the removal of both critics, indicating that the network is generally capable of reproducing the correct words from the corresponding video samples while relying only on the three proposed L1 losses. 

\begin{table}
\centering 
\def\arraystretch{1.5}
\begin{tabular}{  m{2.5cm} | >{\centering}m{1cm}  >{\centering}m{1cm}  >{\centering}m{1cm}  >{\centering\arraybackslash}m{1.1cm} } 
\hline
Model & PESQ & STOI & MCD & WER  \\
\hline
\hline
w/o $L_{PASE}$ & 1.44 & 0.520 & 38.19 & 22.66\,\%  \\
\hline
w/o $L_{power}$ & 1.37 & 0.503 & 39.59 & 24.32\,\%  \\
\hline
w/o $L_{MFCC}$ & 1.44 & 0.518 & 39.03 & \textbf{21.70}\,\%  \\
\hline
w/o Waveform Critic, w/o Power Critic & 1.43 & 0.516 & 38.48 & 22.82\,\%  \\
\hline
Full Model & \textbf{1.47} & \textbf{0.523} & \textbf{37.91} & 23.13\,\%\\
\hline
\end{tabular}
\caption{Ablation study performed on GRID for unseen speaker speech reconstruction.}
\label{ablation_unseen}
\end{table}

\subsection{Comparison with Other Works}
We present our comparison with other works \cite{DBLP:journals/corr/abs-1906-06301,DBLP:journals/corr/abs-2004-02541} on the subject-independent split of GRID in Table \ref{grid_unseen_comparison}. It is clear that our model outperforms previous works in all performance measures. Although, the improvement in PESQ and STOI compared to these works is not as emphatic as the gains reported for seen speakers, WER sees a very substantial reduction. This improvement in WER can easily be observed in our synthesized speech, and clearly shows that our model is far more consistent for this task than previous approaches. Furthermore, the observed MCD is substantially lower in our work, indicating that our synthesized speech yields more accurate spectrograms, which suggests a greater similarity between the content of real and synthesized samples.

\begin{table}
\centering 
\def\arraystretch{1.5}
\begin{tabular}{  m{2.5cm} | >{\centering}m{1cm}  >{\centering}m{1cm}  >{\centering}m{1cm}  >{\centering\arraybackslash}m{1.1cm} } 
\hline
Method & PESQ & STOI & MCD & WER \\
\hline
\hline
GAN-based \cite{DBLP:journals/corr/abs-1906-06301} & 1.24 & 0.470 & 51.28 & 37.10\,\%  \\
\hline
Vocoder-based \cite{DBLP:journals/corr/abs-2004-02541} & 1.23 & 0.477 & 55.02 & 55.23\,\%  \\
\hline
Ours & \textbf{1.47} & \textbf{0.523} & \textbf{37.91} & \textbf{23.13}\,\%  \\
\hline
\end{tabular}
\caption{Comparison between our current and previous model, using full GRID (33 speakers) with an unseen speaker split.}
\label{grid_unseen_comparison}
\end{table}
\subsection{Additional Experiments}

\begin{figure*}
    \centering
    \begin{subfigure}{0.3\linewidth}
        \includegraphics[width=\linewidth]{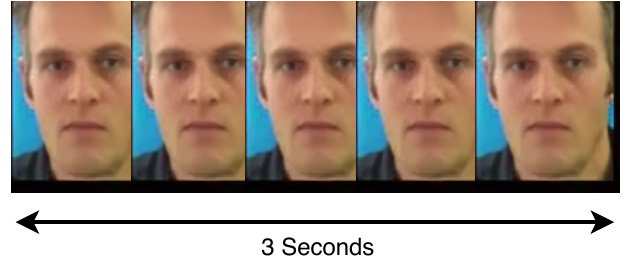}
        \caption{Silent Video}
    \end{subfigure}
    \begin{subfigure}{0.3\linewidth}
        \includegraphics[width=\linewidth]{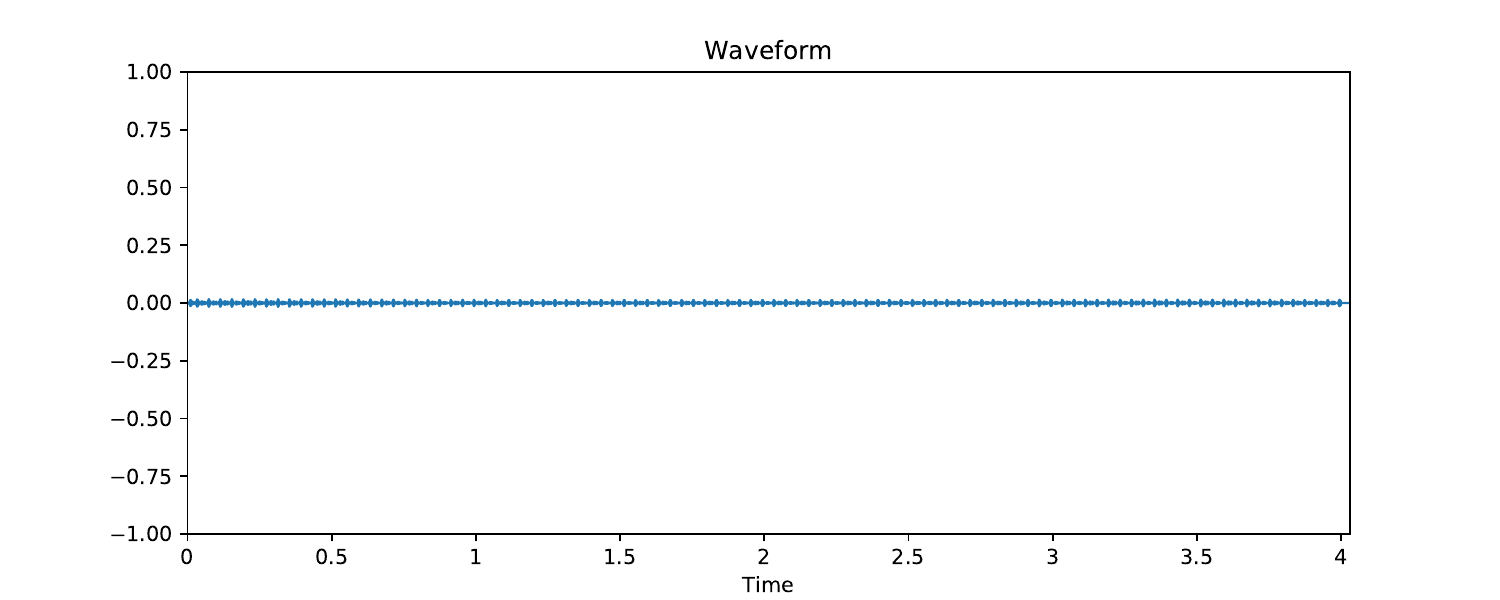}
        \caption{Waveform}
    \end{subfigure}
    \begin{subfigure}{0.3\linewidth}
        \includegraphics[width=\linewidth]{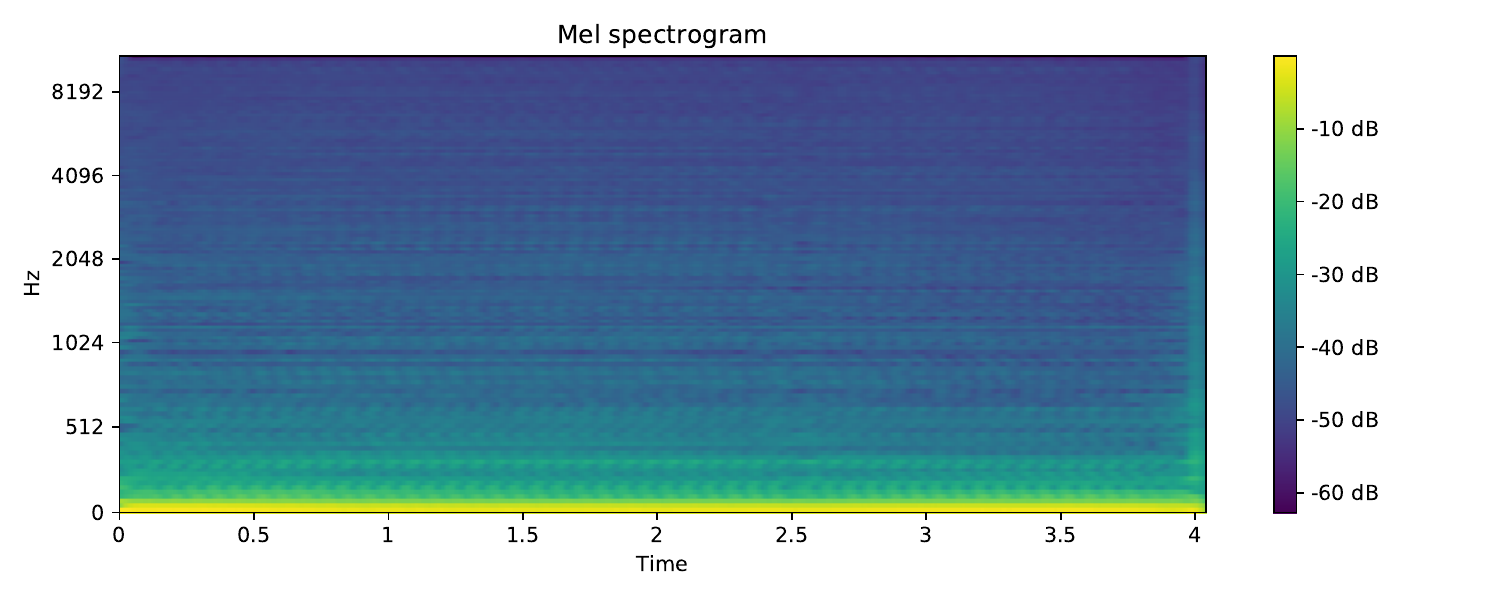}
        \caption{Mel-frequency spectrogram}
    \end{subfigure}
\caption{The spectrogram and waveform for the audio produced by our model for a video of a silent speaker (Speaker 2 from GRID) are portrayed in (a). As displayed in the waveform (b), the audio is almost completely silent, disregarding some low frequency noise which is higlighted in the spectrogram (c). This shows that our model is robust to the scenario of silent speakers and produces minimal baseline noise under these circumstances. This audio sample is also available on our website$^{\ref{note1}}$.}
\label{silent_experiment}
\end{figure*}

Additionally, we present a study on silent speakers. For this experiment, we artificially produce a video of a speaker from the GRID corpus being silent for five seconds by feeding Brownian noise into the facial animation model proposed in \cite{DBLP:journals/ijcv/VougioukasPP20}. We then use this video as input for our model trained on the full GRID dataset (33 speakers, unseen speaker split). This aims to measure two distinct factors: firstly, our model's ability to recognize a silent speaker and not produce any voiced speech; and secondly, the baseline noise that is present in the audio we synthesize with our network, which is clear to observe when the speaker is silent. As discussed in Figure \ref{silent_experiment}, our model performs well in this scenario and produces minimal noise for this silent example.

\section{Results in the Wild} \label{section6c}
In this section, we investigate the performance of the proposed approach on utterances recorded `in the wild'. For this purpose, we use the full LRW dataset, and its subsets FLRW 500 Words, FLRW 100 Words and FLRW 20 Words, which are introduced in Section \ref{section4}. We split the utterances using the default split for LRW (90-5-5\,\% ratio), such that there is no overlap between the utterances in the training, validation and test sets.
To measure the Word Error Rate (WER) for our samples, we use a pre-trained model (based on \cite{DBLP:conf/icassp/PetridisSMCTP18}) which was trained and tested on full LRW using the same split, and achieve a baseline WER of  1.68\,\% on the test set.
\subsection{Comparison with Other Works}
Our comparison with Lip2Wav \cite{DBLP:journals/corr/abs-2005-08209} on LRW (500 Words) is presented in Table \ref{lrw_comparison}. We compare our model to Lip2Wav on LRW (500 Words), in order to compare our model's performance ``in the wild'' to this recent work. Our work shows a great improvement in PESQ compared to Lip2Wav, which suggests that our samples are able to achieve a superior clarity in this regard. On the other hand, our STOI is very similar to the one reported in Lip2Wav, achieving a slight edge which could indicate a minor improvement in intelligibility.   

\begin{table}
\centering 
\def\arraystretch{1.5}
\begin{minipage}{\linewidth}
\begin{tabular}{  m{2.5cm} | >{\centering}m{1cm}  >{\centering}m{1cm}  >{\centering}m{1cm}  >{\centering\arraybackslash}m{1.1cm} } 
\hline
Method & PESQ & STOI & MCD & WER \\
\hline
\hline
Lip2Wav \cite{DBLP:journals/corr/abs-2005-08209} & 1.20 & 0.543 & - & \textbf{34.20}\footnote{Reported using Google STT API.}\,\%  \\
\hline
Ours & \textbf{1.45} & \textbf{0.556} & \textbf{39.32} & 42.51\,\%  \\
\hline
\end{tabular}
\end{minipage}
\caption{Comparison between our model and Lip2Wav, using the full LRW dataset.}
\label{lrw_comparison}
\end{table}

\subsection{Performance for Different Subsets}
In order to demonstrate our model's ability to reconstruct speech in less constrained conditions, we experiment with the LRW dataset, as well as some of its subsets. These subsets present increasing degrees of challenge, culminating with the full LRW dataset which presents the greatest challenge given its large vocabulary and large variance in video perspective.

Regarding the experiments with frontal LRW, we observe that our model maintains a similar overall quality of outputs for larger vocabularies, as demonstrated by the consistency in PESQ, STOI and MCD. However, it is clear that the more difficult task presented by larger vocabularies yields a decrease in the average accuracy of samples, shown by the increasing WER. This implies that our model scales well with larger datasets, but has difficulties in adapting to larger vocabularies in very unconstrained and inconsistent environments. Even still, the word error rate reported for FLRW 20 Words is noticeably low, implying that our model can realistically reconstruct speech for hundreds of different speakers, even under such `wild' conditions. Finally, we found that the full LRW dataset yields a better performance than our full frontal subset (FLRW 500 Words). Although we expected the frontal data to provide an easier task for the network during training and testing, this result shows that the network benefits strongly from a larger training set, even if the visual data is less consistent.

\begin{table}
\centering 
\def\arraystretch{1.5}
\begin{tabular}{  m{2.5cm} | >{\centering}m{1cm}  >{\centering}m{1cm}  >{\centering}m{1cm}  >{\centering\arraybackslash}m{1.1cm} } 
\hline
Corpus & PESQ & STOI & MCD & WER  \\
\hline
\hline
FLRW 20 Words & 1.43 & 0.523 & 43.87 & 25.00\,\%  \\
\hline
FLRW 100 Words & 1.40 & 0.528 & 41.56 & 36.54\,\%  \\
\hline
FLRW 500 Words & 1.44 & 0.555 & 39.72 & 44.28\,\%  \\
\hline
LRW 500 Words & 1.45 & 0.556 & 39.32 & 42.51\,\%  \\
\hline
\end{tabular}
\caption{Study on the performance of our speech reconstruction model for the three subsets of LRW mentioned in Section \ref{section4}, as well as the full LRW dataset.}
\label{lrw_results}
\end{table}

\section{Conclusion} \label{section7}
In this work, we have presented our end-to-end video-to-waveform synthesis model using a generative adversarial network with two critics on waveform and spectrogram. First, we showed through an ablation study on GRID that the use of our losses, adversarial critics and other choices in training methodology provide a positive impact on the quality of our results for both seen and unseen speaker video-to-speech. Furthermore, we demonstrated through our experiments on LRW that our model is able to generate intelligible speech for videos recorded entirely in the wild by hundreds of different speakers. Finally, we compared our model to previous video-to-speech models and found that it produces the best results on most metrics for GRID and LRW and achieves state-of-the-art performance on PESQ for TCD-TIMIT.

We observed that the choice of good critics as well as adequate comparative losses is fundamental towards obtaining realistic results. Therefore, we believe that the pursuit of alternative loss functions (including different adversarial losses) is a promising option for future work. Additionally, we believe that there would be substantial benefit in experimenting with a speaker embedding as input to the generator, in addition to the video, in order to generalize to unseen speakers with a more accurate voice profile, as proposed in \cite{DBLP:conf/icassp/ShenPWSJYCZWRSA18,DBLP:journals/corr/abs-2005-08209}. Finally, extending our model towards other practical applications such as speech inpainting \ie reconstructing missing audio segments in an audiovisual stream, would be a promising research pursuit in order to show the empirical value of video-to-speech synthesis.  

\section*{Acknowledgments}
All datasets used in the experiments and all training, testing and
ablation studies have been contacted at Imperial College. Rodrigo Mira
would like to thank Samsung for their continued support of his work on
this project. Additionally, the authors would like to than AWS for
providing cloud computation resources for the experiments discussed in
this paper.

\ifCLASSOPTIONcaptionsoff
  \newpage
\fi

\section*{References}
\AtNextBibliography{\footnotesize}
\printbibliography[heading=none]

\end{document}